\documentclass{article}

\usepackage[a4paper, total={6in, 9in}]{geometry}
\usepackage{natbib}

\usepackage[utf8]{inputenc} 
\usepackage[T1]{fontenc}    
\usepackage{hyperref}       
\usepackage{booktabs}       
\usepackage{amsfonts}       
\usepackage{nicefrac}       
\usepackage{microtype}      
\usepackage{url}            
\hypersetup{colorlinks, citecolor=BlueViolet, linkcolor=Red, urlcolor=Blue}

\usepackage{mathtools}

\usepackage[dvipsnames]{xcolor}

\usepackage{amsmath}
\usepackage{multirow}
\usepackage{graphicx}
\usepackage[dvipsnames]{xcolor}
\usepackage{epstopdf}
\usepackage{caption}
\usepackage{subcaption}
\usepackage{bm}
\usepackage{pgfplots}
\usepackage{footnote}
\usepackage{stfloats}
\usepackage{placeins}
\usepackage{color,soul}
\usepackage{dsfont}
\usepackage{amssymb}
\usepackage[ruled,vlined]{algorithm2e}
\usepackage{pbox}
\usepackage{enumerate}
\usepackage{etextools}
\usepackage{amsthm}
\usepackage{tabulary}
\usepackage{dsfont}

\usepackage{wrapfig}

\usepackage{float}

\usepackage{cleveref}
\usepackage{comment}

\usepackage{wrapfig}

\def\b{{\text{\textbf{\textit{b}}}}}

\def\sj{{ j^*}}

\def\x{{\mathbf x}}

\def\v{{\mathbf v}}
\def\d{{\mathbf d}}
\def\a{{\mathbf a}}

\def\D{{\mathbf D}}

\def\A{{\mathbf A}}

\def\alfa{{\boldsymbol \alpha}}

\def\gama{{\boldsymbol \gamma}}

\def\bs{{ \backslash}}

\newcommand{\abs}[1]{{\lvert #1 \rvert}}

\newcommand{\argmin}{\mathop{\rm argmin}}
\newcommand{\argmax}{\mathop{\rm argmax}}

\usepackage{thmtools, thm-restate}
\newtheorem{theorem}{Theorem}[section]
\newtheorem{lemma}[theorem]{Lemma}

\def\equationautorefname~#1\null{(#1)\null}

\usepackage{xr}
\makeatletter
\newcommand*{\addFileDependency}[1]{
  \typeout{(#1)}
  \@addtofilelist{#1}
  \IfFileExists{#1}{}{\typeout{No file #1.}}
}
\makeatother

\title{Recovery and Generalization in Over-Realized\\ Dictionary Learning}

\begin{document}

\author{Jeremias Sulam\\ Johns Hopkins University \\ \texttt{jsulam1@jhu.edu} 
\and 
Chong You \\ University of California, Berkeley \\ \texttt{cyou@berkeley.edu}
\and 
Zhihui Zhu \\ Denver University \\ \texttt{zhihui.zhu@du.edu} 
}

\date{}
\maketitle

\begin{abstract}
In over two decades of research, the field of dictionary learning has gathered a large collection of successful applications, and theoretical guarantees for model recovery are known only whenever optimization is carried out in the \emph{same} model class as that of the underlying dictionary. 
This work characterizes the surprising phenomenon that dictionary recovery can be facilitated by searching over the space of larger \emph{over-realized} models. 
This observation is general and independent of the specific dictionary learning algorithm used. We thoroughly demonstrate this observation in practice and provide an analysis of this phenomenon by tying recovery measures to generalization bounds. { In particular, we show that \emph{model recovery} can be upper-bounded by the empirical risk, a model-dependent quantity and the generalization gap, reflecting our empirical findings.} We further show that an efficient and provably correct distillation approach can be employed to recover the correct atoms from the over-realized model. As a result, our meta-algorithm provides dictionary estimates with consistently better recovery of the ground-truth model.
\end{abstract}


\section{Introduction}

Latent variable models have been very successful for a variety of unsupervised learning problems, from regularizing inverse problems of different kinds to enabling clustering, classification or other down-stream supervised learning problems \citep{bengio2013representation}. We focus on sparse representation models, which posit that data $\x\in \mathcal{X} \subseteq \mathbb{R}^d$ admits a sparse decomposition in terms of a redundant dictionary $\D\in \mathcal{D} \subset \mathbb R^{d\times p}$, where $p>d$ and $\mathcal{D}$ is an appropriate constraint set. In other words, $\x = \D\gama$, where the 
number of nonzero entries is small: $\| \gama \|_0 \leq k \ll d$. These models are most useful when the model $\D$ is learned from a collection of samples $\{\x_i\}^n_{i=1}$, thus allowing for greater sparsity or representation power. This task goes by the name of \emph{dictionary learning}, and many algorithms have been proposed over the last two decades to (most often approximately) solve this problem \citep{aharon2006ksvd,mairal2010online,engan1999method,olshausen1997sparse,arora2015simple}.

A central problem in dictionary learning is that of model recovery. More precisely, assuming that the training samples follow such a generative model, $\x_i = \D\gama_i$, and one has access to a learning algorithm that provides an estimate $\hat{\D}$, how close will the obtained model be from the true generating dictionary? There exist by now a rich literature on these questions. Some of these results are concerned with providing recovery guarantees for popular and practical dictionary learning methods, such as the K-SVD \citep{aharon2006uniqueness,schnass2014identifiability} or simpler online learning algorithms \citep{olshausen1997sparse,arora2015simple}. Others instead propose new algorithms with recovery guarantees, most often in an alternating minimization manner \citep{agarwal2016learning,agarwal2014learning,arora2014more,arora2014new,arora2017provable}, while other results study local identifiability \citep{geng2014local,gribonval2015sparse} or fundamental limits and min-max optimal bounds \citep{shakeri2018minimax,jung2016minimax}. Naturally, these guarantees depend on the minimum number of training samples, $n$, as well as on the parameters of the model: $d,p$ and $k$, the particular distribution of the non-zero values, and possibly the amount of noise contamination in the observations.

Though dictionary learning algorithms vary, by and large they share the following common scheme: given the constraint set $\mathcal{D}_p$ of the ground-truth model, typically $\mathcal D_p = \{ \D \in \mathbb R^{d\times p} : \|\D_i\|_2 = 1~ \forall i \in \{1,\ldots,p\} \}$, and given a collection of $n$ samples from this model, one searches for an estimate $\hat{\D} \in \mathcal{D}_p$ by means of some optimization approach. The first question we pose in this work is the following: \emph{Why should one limit to the set $\mathcal{D}_p$ instead of searching over a larger class of models?} Somewhat surprisingly, we will show that dictionary recovery can be consistently improved if one allows the learning algorithm to search for models $\tilde{\D}\in {\mathcal{D}_{p'}}\subset \mathbb R^{d\times p'}$, where $p'>p$. In other words, we will search for a larger set of atoms than those that are strictly necessary to sparsely represent the training data -- an \emph{over-realized} model. 

While it is certainly natural that a larger model of $p'>p$ atoms can approximate the training samples better than one with $p$ atoms, it is not immediately obvious that this might lead to a better overall dictionary recovery. After all, how can one evaluate model recovery if the estimate and ground-truth models belong to different spaces? To this end, we propose a new distance metric and show that it can be upper bounded by a function of the empirical risk (i.e. training error) and the generalization gap, both of which are computable. This result links recovery guarantees to generalization bounds, allowing us to characterize the behaviour observed in our experiments, and leading to a uniform upper bound to the recovery error.

{Even if one can improve recovery with a larger model, one might be interested in obtaining a dictionary of the original size, i.e. only with $p$ columns}. We therefore study a second driving question: \emph{given a trained model $\tilde{\D}\in{\mathcal{D}_{p'}}$, can one distill from it an estimate $\hat{\D}\in\mathcal{D}_p$ and, in doing so, improve the recovery of the true dictionary?} We will answer this question in the affirmative, providing a provably correct algorithm under incoherence assumptions. As a result, we will provide a meta-algorithm for dictionary learning via over-realized models that improves model recovery over conventional (non over-realized) approaches, across a variety of model parameters and learning algorithms.

The study of over-realized models in unsupervised learning has received some -- but limited -- attention in the past. The work by \cite{dasgupta2007probabilistic} showed more than a decade ago that the recovery of $k$ clusters by k-means \citep{lloyd1982least} can be improved by a two-step process, whereby in the first round one uses more random guesses as initialization (more precisely, $\mathcal{O}(k \log{k})$). To the best of our knowledge, the recent inspiring work by \cite{buhai2019benefits} is the first to show empirical benefits of over-realized models in representation learning settings. In this work, the authors demonstrate that over-realization can lead to higher log-likelihood and improved recovery in noisy-OR networks and for a particular dictionary learning algorithm \citep{li2016recovery}. 
In the neural networks community, a new and growing body of work has shown that a large number of parameters is the key to obtaining good empirical performance \citep{zhang2016understanding}, 
bringing forth a surge of interests for providing theoretical support \citep{goldt2019dynamics,tian2019over,mei2019generalization,belkin2019reconciling,yang2020rethinking}. This \emph{over-parameterization} regime refers to models having a larger number of parameters than training samples. In contrasts, in this work we study how \emph{over-realization} (having more parameters than that of the underlying generative model) improves recovery in a dictionary learning.

\paragraph{Overview} We first introduce our notation and provide the necessary background in \autoref{sec:Preliminaries}. We then address the recovery problem in the over-realized case in \autoref{sec:seraching_overrealized}, providing examples and presenting our main theoretical result. \autoref{sec:distill} tackles the question of the distillation of  larger models, and provides a provably correct algorithm as well as extensive empirical evidence. We finally delineate final remarks and conclude in \autoref{sec:conclusion}.

\section{Preliminaries}
\label{sec:Preliminaries}

We consider data $\x\in \mathbb{R}^d$, and a redundant dictionary $\D_0\in \mathcal D_p$, $p>d$. We consider the following generative model for $\x$ throughout this work, providing a sampling distribution $\mathbb P$: a sparse representation $\gama \in\mathbb{R}^p$ is sampled from a set of $k$-sparse vectors by (i) sampling its support $S$ uniformly from the set of all possible $\binom{p}{k}$ supports of cardinality $k$, and (ii) sampling its non-zero values i.i.d from a distribution $\mathcal{P}$, $\gama_i \sim \mathcal{P}~\forall i\in S$ with mean zero and unit variance (for simplicity). Samples are then obtained as $\x=\D_0\gama$. Given $\x$ and $\D_0$, the problem of retrieving the representation $\gama$ is termed \emph{sparse coding}, and it involves solving a problem of the form
\begin{equation}\label{eq:pursuit}
    \min_{\gama } \frac{1}{2}\| \x - \D_0\gama\|^2_2 + g(\gama),
\end{equation}
where $g(\gama)$ is a sparsity-promoting function that regularizes the ill-posed recovery problem. Typical choices for $g$ are the non-convex and non-smooth $\ell_0$ pseudo-norm, or its convex relaxation, the $\ell_1$ norm. Alternatively, $g$ may denote an indicator function over a constraint set, such as
\begin{equation} \label{eq:indicator_set}
    g_k(\gama) = \left\{
	\begin{array}{ll}
		0  & \mbox{if } \|\gama\|_0 \leq k, \\
		+\infty & \mbox{otherwise.}
	\end{array}
\right.
\end{equation}
In either case, numerous pursuit algorithms exist that allow for the provable recovery of $\gama$ under assumptions like restricted isometry property \citep{candes2005decoding} or incoherence \citep{tropp2004greed,donoho2003optimally}. These exact recovery guarantees are naturally extended to approximate recovery in the case of noisy measurements. When $g(\gama) = \|\gama\|_1$, the problem is termed Basis Pursuit DeNoising or Lasso \citep{tibshirani1996regression} (and Basis Pursuit when an $\ell_1$ ball is used as a constrained set). Alternatively, one may employ greedy algorithms such as the popular Orthogonal Matching Pursuit (OMP) \citep{pati1993orthogonal}, which approximates the solution to the $\ell_0$-constrained problem.

When the dictionary is not known, the dictionary learning problem attempts to recover an estimate as close as possible to the ground-truth model given a set of $n$ training samples $\x_i$ from it.
The quality of a dictionary in approximating a sample $\x$ is measured by the function value of the cost above, namely
\begin{equation}
f_\x(\D) \coloneqq \inf_{\gama \in \mathbb R^{p}} \frac{1}{2}\| \x - \D\gama\|^2_2 + g(\gama).
\end{equation}
In this way, the dictionary learning problem minimizes this loss over the $n$ samples, and can be written as
\begin{equation}\label{eq:dict_Learning}
    \min_{\D\in\mathcal D_p} \frac{1}{n} \sum^n_{i=1} f_{\x_i}(\D).
\end{equation}
The resulting optimization problem is non-convex and hard to analyze in general \citep{tillmann2014computational}, but this has not prevented the development of many  -- and very successful --  algorithms. One such methods is the Online Dictionary Learning (ODL) from \cite{mairal2010online}, which minimizes \eqref{eq:dict_Learning} in an online manner. In a nutshell, given a current estimate for the dictionary, this algorithm iterates between drawing a sample (or a mini-batch thereof) at random, then employing a pursuit algorithm to minimize \eqref{eq:pursuit}, and finally updating the dictionary so as to minimize a surrogate of the cost in \eqref{eq:dict_Learning}. The approach is general in that it can accommodate different pursuit algorithms for different penalty functions $g(\gama)$, and it scales well to large datasets. The very popular K-SVD \citep{aharon2006ksvd}, on the other hand, is a batch-learning approach that alternates between sparse coding (typically with OMP) and dictionary update, which is characteristically carried out column-by-column by performing rank-1 approximations to atom-wise residual.

\paragraph{Recovery}
A central question is this setting is that of model recovery, which studies how far the recovered estimate $\hat{\D}\in\mathcal D_p$ is from the ground-truth dictionary, $\D_0\in\mathcal D_p$. To formalize this question one needs an appropriate measure of distance between matrices. The problem in \eqref{eq:dict_Learning} is permutation (and sign) invariant: the columns of the dictionary can be arbitrarily permuted (or multiplied by $-1$) without modifying the cost $f_\x(\D)$. Thus, different measures of recovery have been used in previous works accounting for such invariance, such as \citep{arora2015simple}
\begin{equation} \label{eq:old_distance_def}
    \min_{P \in \Pi}\|\D_0 - \hat\D P \|_F^2,
\end{equation}
where $\Pi$ is the set of signed permutation matrices, i.e orthogonal matrices that contain only $\{0, \pm 1\}$. Several works have addressed these questions of recovery over the last decade. Some of these show local linear convergence to the global optimum (i.e. the true model) via alternating minimization employing $\ell_1$ penalty functions \citep{agarwal2014learning,agarwal2016learning} or to an $\epsilon$-close optimum via $\ell_0$ constraints \citep{arora2015simple}. In the simpler case of orthonormal dictionaries the optimization landscape is better understood \citep{zhai2019complete}, as in the case of learning only one atom  \citep{sun2015nonconvex,qu2019geometric}. In these settings, these non-convex problems have a benign geometry structure that allows for provable algorithms. On the other hand, \citep{jung2016minimax} develops minimax risk bounds for dictionary recovery, and \citep{shakeri2018minimax} studies these as a function of their tensor structure. All of these result, however, analyze the conventional setting whereby the constraint sets of the ground-truth dictionary and the one enforced during optimization are the same.

\paragraph{Generalization Gap}

From a statistical learning standpoint, the dictionary learning problem consists in finding a model $\hat{\D} \in \mathcal{D}_p$ that minimizes the above function in expectation over the population, i.e.,
\begin{equation} \label{eq:learning_problem}
    \hat{\D} \in \argmin_{\D\in\mathcal{D}_p} ~~ \underset{\x\sim\mathbb P}{\mathbb E} ~ \left[ f_\x(\D) \right].
\end{equation}
Since one does not typically have access to the underlying distribution, the empirical risk minimization algorithm (ERM) minimizes the empirical estimate of the above risk, which is precisely the problem in Eq. \eqref{eq:dict_Learning}. In this context, a central question is given by the generalization gap, which quantifies the extent to which the empirical error, $\mathcal{R}_S(\D)=\frac{1}{n}\sum_{i=1}^n f_{\x_i}(\D)$, differs from its expectation in Eq. \eqref{eq:learning_problem}, termed generalization error, or risk. Uniform bounds have recently been developed for these models \citep{maurer2010k,vainsencher2011sample,seibert2019sample}. More specifically, with overwhelming probability over the draw of the samples, the work in \cite{gribonval2015sample} shows that this is uniformly bounded,
\begin{equation}\label{eq:generalization-gap}
    \sup_{\D\in\mathcal{D}_p} \left| \mathcal{R}_S(\D) - \underset{\x\sim\mathbb{P}}{\mathbb{E}} [ f_\x(\D) ] \right| \leq \eta_n,
\end{equation}
where $\eta_n$, depends on the model {capacity}, the number of samples, as well as the data distribution and properties     of the penalty function $g$. Slightly more specifically, $\eta_n$ is $\mathcal{O}(\sqrt{(dp)\log{n}/n})$, where $(dp)$ is the number of parameters in the dictionary with $p$ atoms. This type of bounds are very useful, since they provide an upper bound to the expected (real) risk given the empirical risk, and they reflect the natural trade-off between the model size (number of atoms, $p$) and the number of training samples, $n$. The bound above holds not just for norms and \emph{norm-like} regularization functions (like the $\ell_1$ norm) but also for indicator sets as $g_k$ in \eqref{eq:indicator_set}. We will keep our derivations maximally general by simply referring to $\eta_n$, and we refer the reader to \citep{gribonval2015sample} for further details on the involved constants.

\section{Searching for over-realized dictionaries}
\label{sec:seraching_overrealized}

\begin{figure}
\centering
\subcaptionbox{\label{fig:Risk_1}}
{\includegraphics[width = .24\textwidth,trim = 5 0 5 0]{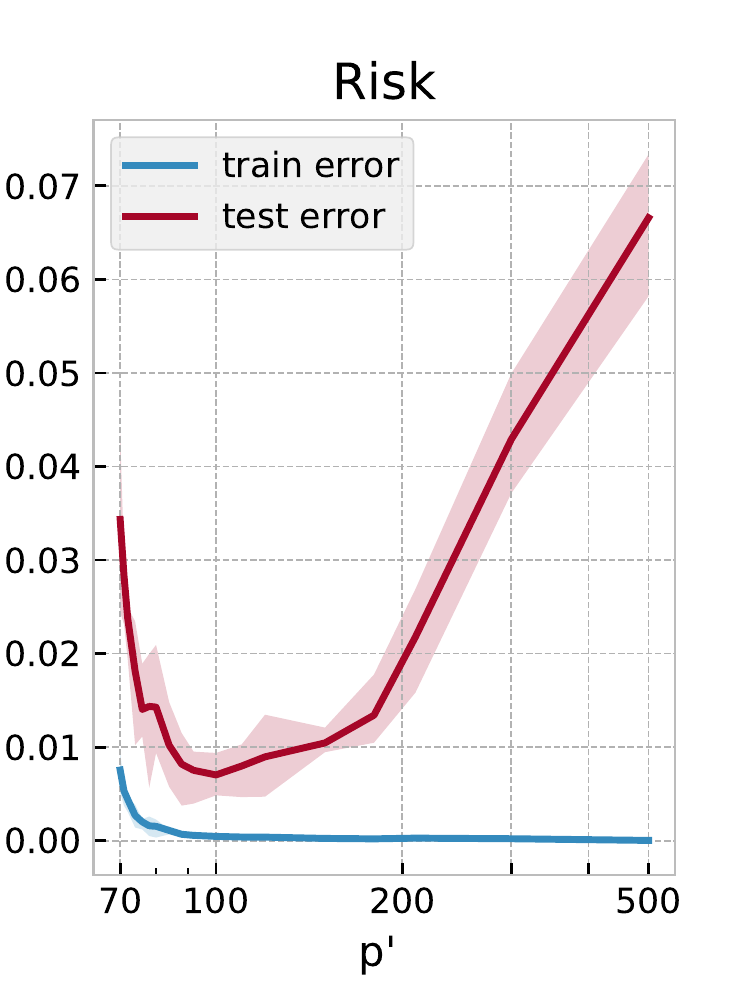}}
\subcaptionbox{\label{fig:DictDist_1}}
{\includegraphics[width = .24\textwidth,trim = 5 0 5 0]{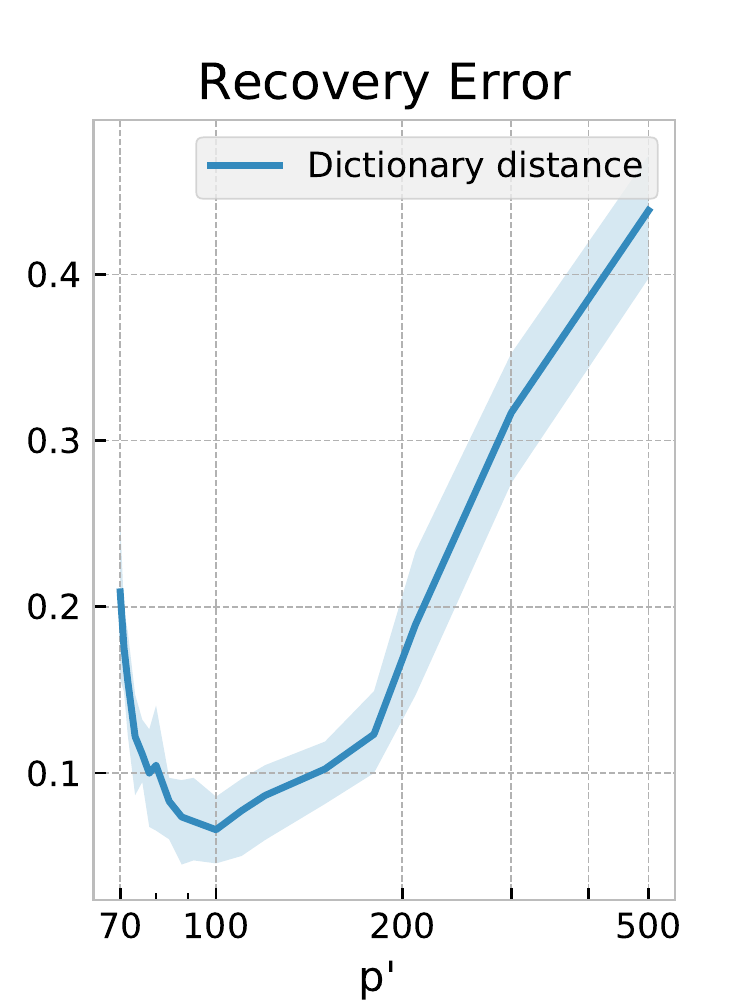}}
\subcaptionbox{\label{fig:Risk_varyingData}}
{\includegraphics[width = .24\textwidth,trim = 5 0 5 0]{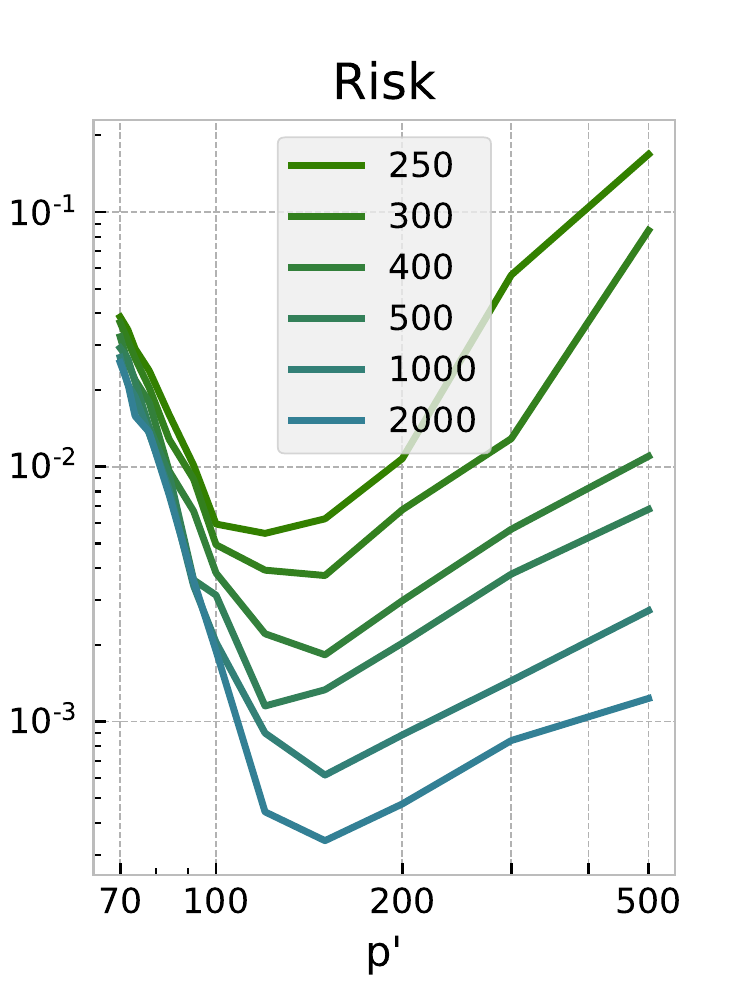}}
\subcaptionbox{\label{fig:DictError_varyingData}}
{\includegraphics[width = .24\textwidth,trim = 5 0 5 0]{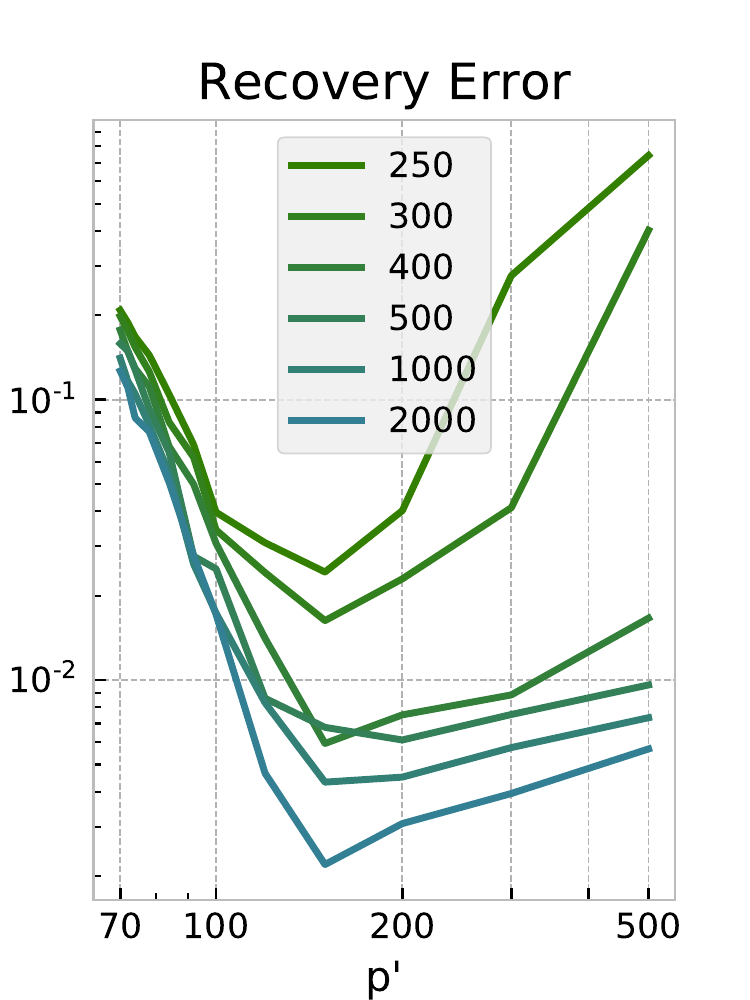}}
\caption{ (a) and (b): Risk of the estimated dictionary and distance to the ground truth model, as defined in Eq. \eqref{eq:distance_measure}, trained with 300 samples. (c) and (d): Risk (test error) and recovery error for different size of the training data. The dictionary size $p'$ refers to that of the estimated matrix, whereas the original one remains fixed containing $p=$ 70 atoms.}
\label{fig:Exper_1}
\end{figure}

In this work we focus on the over-realized setting, in which the minimization in Eq. \eqref{eq:dict_Learning} is done over a class of dictionaries $\mathcal D_{p'}$, with $p'>p$, i.e. larger than the original model. {One might wonder as to the need for this change. After all, there exists indeed a global minimum ($\D_0$) with $p$ atoms that achieves both zero training and testing errors. Nonetheless, one should keep in mind that the optimization landscape of these optimization problems is still not fully understood, and practical local-search algorithms may converge to a local minimum due to the high non-convexity of the problem \eqref{eq:dict_Learning}.}

We first require a distance measure between dictionaries\footnote{We will use $\D^0_i$ to denote the $i^{th}$ column, or atom, from $\D_0$.} of potentially different sizes. We will use the following definition for the distance between a dictionary $\D_0\in\mathcal D_p$ and an estimate $\hat{\D}\in\mathcal D_{p'}$:
\begin{equation} \label{eq:distance_measure}
    d(\D_0,\hat\D) \coloneqq \frac{1}{p} \sum_{i=1}^p \min_{j\in[p']} \min_{c\in\{-1,1\}} \|\D^0_i - c~ \hat{\D}_j \|^2_2.
\end{equation}
Note that this distance is zero if and only if there exists a match for each of the atoms in $\D_0$ in the estimated $\hat{\D}$, irrespective the size $p'$. Moreover, this expression provides a generalization\footnote{Note that our definition in Eq. \eqref{eq:distance_measure} generalizes that in Eq. \eqref{eq:old_distance_def} by allowing the set of permutation matrices to become column-selection (non-square) ones.} of the commonly used distance measure in \eqref{eq:old_distance_def}. 

We now explore the first question posed above, namely: can one obtain an estimate with better generalization error \emph{and} lower recovery error by searching in a hypothesis class bigger than that of the original dictionary? As a motivating example, we construct the following experimental setting. Data is sampled as described in the previous section from a ground truth dictionary (with normalized Gaussian atoms) of size $50\times 70$, from representations with cardinality $k=3$. We construct 300 such samples for training, leaving 1000 to estimate the population statistics. As a learning algorithm, we employ ODL\footnote{Available at \url{spams-devel.gforge.inria.fr/}. Note that ODL can accommodate different formulations and algorithms for the sparse coding step (and not just an $\ell_1$ minimization), which will enable us to explore different experimental settings.} \citep{mairal2010online} for 2000 iterations, which are more than sufficient for convergence. We employ OMP for the sparse coding step.

In \autoref{fig:Risk_1} we depict the risk, or error, on both training and test sets, as a function of the number of atoms in the estimated dictionary $\hat{\D}$, from 70 (the size of the ground-truth model) to 500. We repeat the experiment 20 times, and present the mean together with the $25\%$ and $75\%$ percentiles. Interestingly, both train and testing errors, shown in \autoref{fig:Risk_1}, \emph{improve} with increasing dictionary size $p'>p$ within some range.
More surprisingly, the distance from the estimate to the ground truth $\D_0$ also improves as one searches for bigger dictionaries. Note that because of our definition of distance in Eq. \eqref{eq:distance_measure}, a small distance implies a close recovery of the true atoms, irrespective of the ``extra'' ones. 
At the same time, this behaviour is tightly related to that of model capacity and over-fitting: while increased dictionary size allows for better recovery, the finite training size eventually becomes insufficient to train the larger model and the generalization error increases (while perfectly fitting the training data). This is verified in  \autoref{fig:Risk_varyingData} and \autoref{fig:DictError_varyingData}, seeing that the generalization error --  \emph{and dictionary recovery} -- is precisely controlled by the size of the training set. In this figure, only the means of the 20 realizations are depicted for the sake of clarity.


\subsection{Recovery guarantees via generalization bounds}
While the behaviour observed in \autoref{fig:Risk_1} and \autoref{fig:Risk_varyingData} is well understood in the statistical learning literature, {this is still surprising in light of the fact that \emph{there exist} a ground truth model with just $p$ atoms that achieves zero risk.}
Moreover, how this relates to improved recovery of the ground-truth dictionary in over-realized settings -- as shown in  \autoref{fig:DictDist_1} and \autoref{fig:DictError_varyingData} -- is, to the best of our knowledge, unknown. Learning bounds and recovery guarantees for dictionary learning have so far remain mostly separated. We will now precisely connect the model recovery error with its expected risk, providing a theoretical characterization for this phenomenon.

Let $f^{[s]}_{\x_i}(\hat{\D}) = \inf_{\gama: \|\gama\|_0\le s} \frac{1}{2}\| \x - \hat{\D}\gama\|^2_2$ denote the loss measured with $s$ non-zero coefficient. We will denote the mutual coherence of a dictionary by $\mu(\D) = \max_{i\neq j} | \langle \D_i , \D_j \rangle|$ (recall that columns are normalized).
Furthermore, for a given atom in the estimate dictionary, $\hat{\D}_j$, consider its closest atom in the ground truth dictionary, $\D^0_{i_{(j)}}${, where $i_{(j)} = \argmin_{i \in [p]} \min_{c \in \{-1, 1\}} \|\D_i^0 - c \hat\D_j\|_2$}. 
We will also need a cross-dictionary coherence, defined as 
\[\nu{(\hat\D, \D_0)} = \max_j \max_{k\neq i_{(j)}} \left| \langle \hat\D_j , \D^0_k \rangle \right|.\]
In words, $\nu(\hat\D, \D_0)$ quantifies the coherence between $\hat{\D}$ and $\D_0$ after excluding the closest neighbor of each atom. While this expression might seem somewhat convoluted, this simply reduces to the traditional mutual coherence of the dictionary, $\mu(\D_0)$, in the case that $\hat\D = \D_0$. 
With these definitions, we have the following central Lemma.

\begin{lemma}
\label{lemma:distance_bound}

For a ground-truth dictionary $\D_0\in\mathcal D_p$ generating samples $\x_i=\D_0\gama_i$, where $\gama_i$ are $k$-sparse with non-zeros sampled iid from a zero mean and unit variance distribution, and for any dictionary $\hat\D\in\mathcal D_{p'}$, we have that
\begin{equation}
   \frac{2}{k}~ \underset{\x\sim\mathbb P}{\mathbb{E}}~[f^{[k]}_{\x}({{\hat\D}})] ~ \leq ~
d(\hat{\D},\D_0) ~ \leq ~ \frac{4}{k} ~ \underset{\x\sim\mathbb P}{\mathbb{E}}~[ f^{[1]}_{\x}({{\hat\D}})]  - \frac{4}{k}\zeta_k (k-1).
\end{equation}
where $\zeta_k = \max\left\{0~,~ 1-(k-2)\mu(\D_0) - 2\nu(\hat\D, \D_0)^2 \right\}$.
\end{lemma}

Note that this result links the recovery distance, $d(\hat\D,\D_0)$, with the expected risk, as measured by $f^{[1]}_\x(\hat\D)$ and $f^{[k]}_\x(\hat\D)$. We will comment on further implications of this shortly, but first we present our main result as a consequence the \autoref{lemma:distance_bound}, which is of practical relevance. Employing the generalization bound from Eq. \eqref{eq:generalization-gap}, we can bound the dictionary distance by informative quantities, as presented in the main result.

\begin{theorem} \label{thm:distance_bound}
For a ground-truth dictionary $\D_0\in\mathcal D_p$ generating samples $\x_i$ with sparsity of $k$, and for any estimate $\hat\D\in\mathcal D_{p'}$, with overwhelming probability, we have that  \vspace{-.15cm}
\begin{equation}
   \frac{k}{4} d(\hat{\D},\D_0) \leq \frac{1}{n} \sum^n_{i} f^{[1]}_{\x_i}(\hat{\D}) - \zeta_k (k-1) + \mathcal{O}\left(\sqrt{\frac{dp'\log(n)}{n}}\right).
\end{equation}
\end{theorem} 
First, this result shows that the distance to the true model can be upper bounded by the empirical risk up to the generalization gap and a model-dependent quantity. This reflects an important implicit trade-off: dictionary recovery can be decreased by increasing the model capacity (dictionary size) as long as the generalization gap is kept small by increasing the sample size appropriately. This is precisely the behaviour observed in \autoref{fig:DictError_varyingData} above.
Second, the term $\zeta_k(k-1)$ appearing in both results above accounts for the fact that the upper bound is constructed via $f^{[1]}_{\x_i}$, as opposed to $f^{[k]}_{\x_i}$. Indeed, note that this term vanishes when $k=1$. When $k>1$, the empirical estimate of $f^{[1]}_{\x_i}$ will necessarily be greater than zero. It is in these cases where the term $\zeta_k(k-1)$ provides a non-trivial tighter bound, as long as $k\leq 2 + 1/\mu(\D_0) - 2 \nu(\hat\D, \D_0)^2 /\mu(\D_0)$, which are mild conditions. {An upper bound given in terms of $f^{[k]}_{\x_i}$ (instead of $f^{[1]}_{\x_i}$) would be more desirable, but it is unclear if this could be obtained.}

While we defer the proof of \autoref{lemma:distance_bound} to Supplement Material \ref{supp:recovery_guarantees}, let us provide a brief proof sketch. The upper and lower bound for $d(\hat\D,\D)$ are obtained independently, though with similar techniques. For the upper bound, we make the observation that the risk $\mathbb E[f_\x^{[1]}(\hat\D)]$ can be expressed analytically in closed form, and can be further decomposed in three terms. Relying on the fact that the non-zero entries are drawn i.i.d with mean zero and unit variance, the expectation one of these vanishes; another term can be lower bounded by $\zeta_k(k-1)$, while the remaining term can be lower bounded by a quantity that is proportional to the dictionary distance $d(\hat\D,\D_0)$. 
The lower bound, on the other hand, is obtained by constructing an analytical (and potentially sub-optimal) solution for the sparse coding problem represented by $f^{[k]}_\x(\hat \D)$ relying on the atoms that are closest to $\D_0$, thus upper bounding this risk. 
A series of algebraic manipulations and the final evaluation of the expectation 
provide the final upper bound on $\mathbb E[f^{[k]}_x(\hat \D)]$ as a function of the distance $d(\D_0,\hat\D)$.

As we see, these results provide an answer in support of learning larger dictionaries, not only to minimize the expected risk but also to obtain estimates with small distance to the ground-truth model. However, a question remains: how can one \emph{distill} the estimated over-realized $\hat\D$ to recover the best $p$ atoms that are the closest to the real model? This is the question we address in the next section.

\section{Distilling the over-realized model}
\label{sec:distill}


In this section, we will first show that the recovered atoms in the over-realized dictionary exhibit two distinct behaviors: any recovered atom is either (very) close to a true atoms in $\D_0$, or is significantly far apart from all atoms in $\D_0$.
We will also show that this clustering behaviour correlates with the atom usage in the estimated model. From this observation, we will then derive a provably correct pruning strategy based on the atom's usage frequency. This distillation approach will recover an estimate, $\hat\D\in\mathcal D_p$, of the original size with a lower recovery error than the traditional (non over-realized) learning approach.

\begin{figure}[t]
\centering
\subcaptionbox{\label{fig:Exper_3}}
{\includegraphics[width = .32\textwidth,trim = 30 20 30 20]{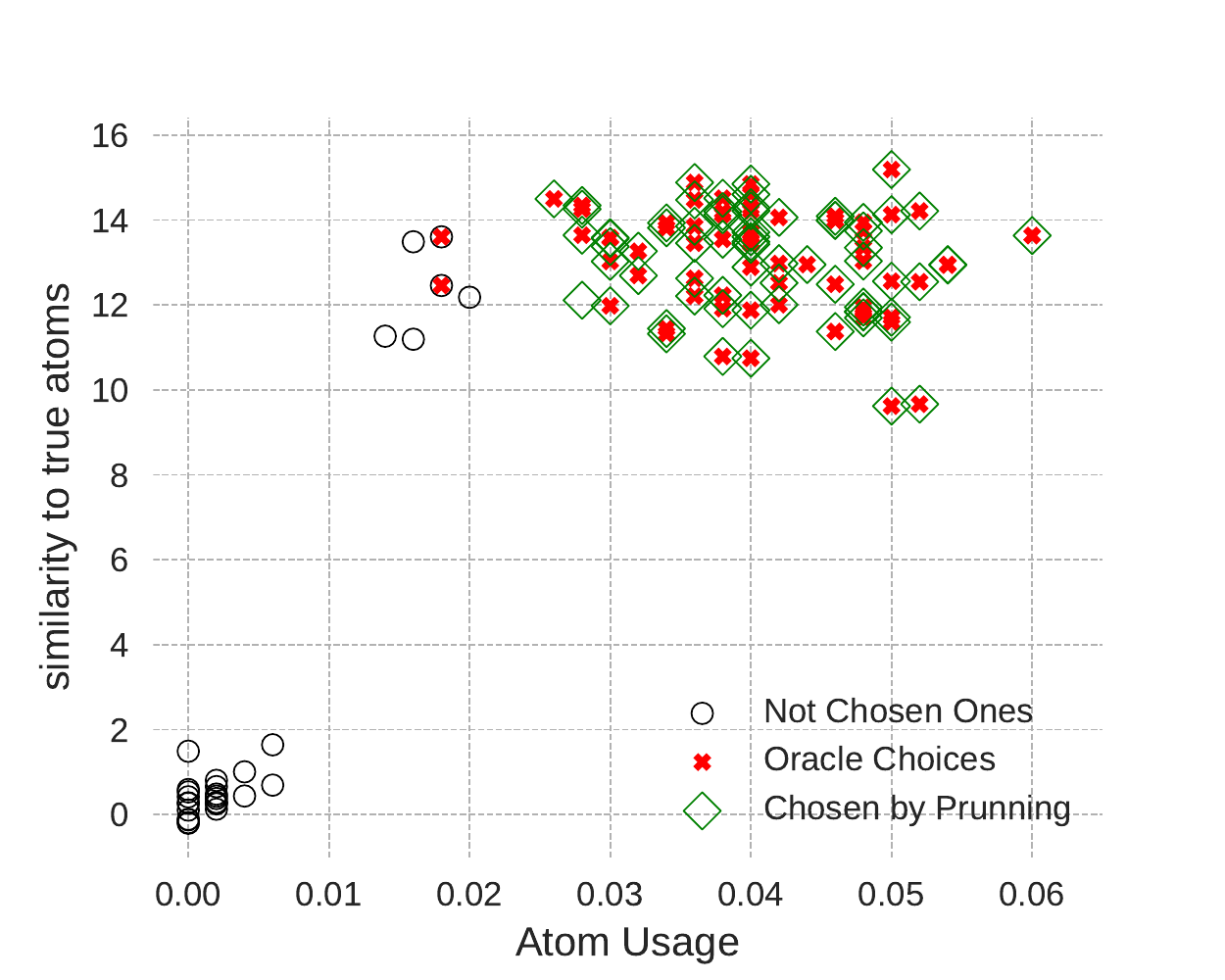}}
\subcaptionbox{\label{fig:prunning_curves}}
{\includegraphics[width = .66\textwidth,trim = 50 20 60 30]{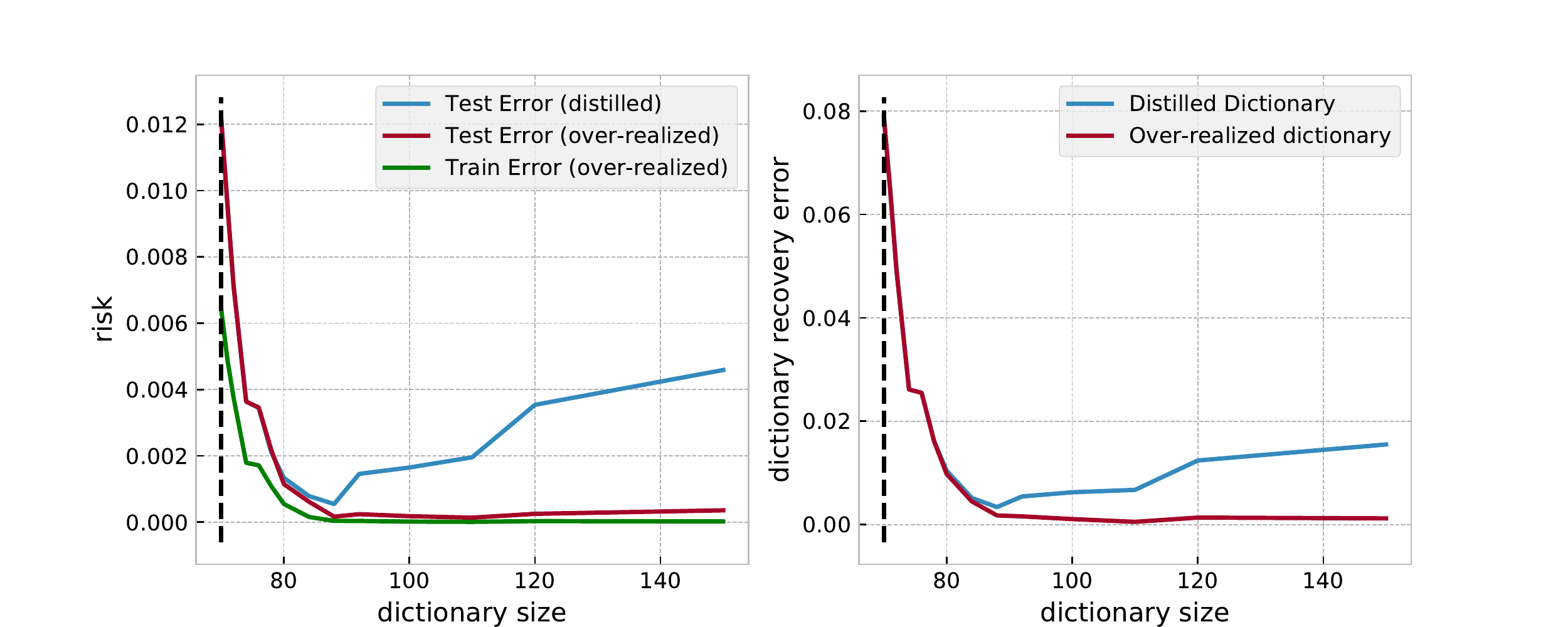}
}
\caption{  (a) Atoms in the over-realized $\hat\D$: their similarity to their closest atom in the ground-truth dictionary $\D$ and its usage frequency. (b) Risk and distance to ground truth model by the over-realized dictionary (i.e. with $p'>p$) and by the distilled version, of the same size as the original model ($p'=p$). }
\label{fig:Exper_4}
\end{figure}
As before, given 500 training samples created as the linear combination of $k=3$ atoms from a ground-truth dictionary $\D_0$ with 70 atoms in 50 dimensions, we train an over-realized dictionary $\hat\D$ with 100 atoms using ODL (with OMP for sparse coding). We then measure, per estimated atom $\hat\D_j$, the similarity to its closest neighbor in the ground-truth $\D_0$ 
{(computed as $-\log{\|\hat\D_j - \D^0_{i_{(j)}}\|^2_2}$).}
We plot these similarities as a function of the atom's \emph{usage}: the relative number of times it is used by the training samples upon completion of training. 
The results are depicted in \autoref{fig:Exper_3}, and two observations are worth noting: the recovered atoms \emph{either} have a high similarity with those in the ground-truth dictionary \emph{or} are markedly distinct, with a clear separation between groups. This is similar to the observation made in \citep{buhai2019benefits} in the context of noisy-or networks and approximate sparse coding. Second, there exists a strong correlation between the former measure -- which cannot be computed in practise, without the original model -- and the number of times an estimated atom is used by the training samples -- which can. 


Following this observation, we then propose the following simple meta-algorithm: after learning an over-realized dictionary, we keep the $p$ most frequently used atoms by the training samples. Other works have suggested similar approaches that prune the over-realized model to a subset of components and then continue the optimization with these  as better initializations \citep{dasgupta2007probabilistic}. This is not needed in our setting, however, likely due to the significant more accurate coding step. \autoref{fig:prunning_curves} illustrates the same experiment as that in  \autoref{fig:Risk_1} and \autoref{fig:DictDist_1}, though now with the statistics provided by our distillation strategy. While clearly the distillation procedure introduces some errors, it still provides a considerable advantage over the traditional approach (i.e. training with the original size $p$) by significantly diminishing the recovery error. This is further explained by the details in \autoref{fig:Exper_3}, comparing the atoms chosen by this distillation procedure and the \emph{oracle} choices -- those atoms that are the closest to the ground-truth dictionary. As can be seen, most atoms selected by this strategy coincide with the oracle ones.

\subsection{Theoretical guarantees for distillation}

We now strengthen our argument for our distillation strategy. In the following result, we show that if the atom usage of the over-realized estimate $\hat\D$ is measured via OMP (with $k=1$), and $\hat\D$ contains at least $p$ atoms that are $\epsilon$-close to the real ones (plus others that are not), then OMP is guaranteed to select the correct (i.e. closest) ones, thus retaining them in the pruning stage.

Let $\D_0 \in \mathbb R^{d\times p}$ and consider, without loss of generality, that $\hat\D = [\hat\D_0, \A] \in \mathbb R^{d\times p'}$, $p'>p$, with $\hat \D_0\in \mathbb R^{d\times m}$, with $p\leq m\leq p'$, such that $d(\hat 
\D_i^0,\D_0) \le \epsilon$ for all $i\in \{1,\ldots,m\}$,
and $d(\A_j,\D^0) > \epsilon$ for all $j\in \{1,\ldots, p'-m\}$. In other words, $\hat \D_0$ contains all those $m$ atoms that are $\epsilon$-close to those in $\D_0$, while $\A$ contains those that are further away. Additionally, we require that each atom in $\D_0$ has at least one $\epsilon$-neighbor in $\hat \D_0$; i.e. $d( 
\D^0_i, \hat \D_0) \le \epsilon$ for all $i\in \{1,\ldots,p\}$. We allow $m\ge p$ since the over-realized estimate $\hat \D$ may naturally contain several atoms that close to a real one. Also suppose that both $\D_0$ and $\hat \D$ are column-wise normalized for simplicity. These assumptions, which reflect the behavior depicted in \autoref{fig:Exper_3}, are illustrated in \autoref{fig:diagram} below. Lastly, let us denote by $\mu(\D_0,\A) = \max_{i,j}\left|\langle  \D_i^0, \A_j \rangle\right|$ the mutual coherence between $\D_0$ and $\A$. 

\begin{figure}[t]
  \begin{center}
     \includegraphics[trim = -10 0 10 0, width=.45\textwidth]{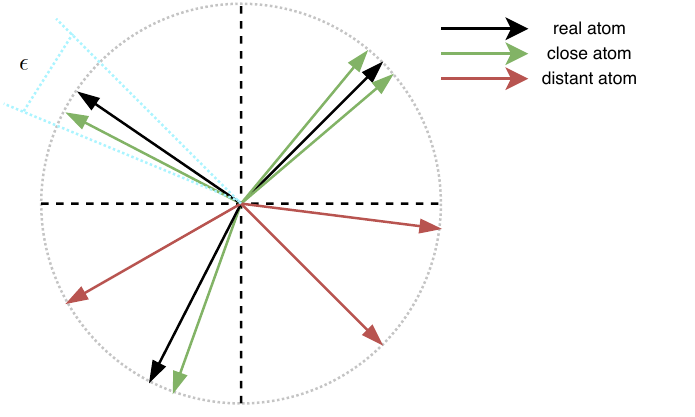}
  \end{center}
  \caption{{Illustration of the true and estimated atoms.}}
  \label{fig:diagram}
\end{figure}

With these definitions, we have the following result, which we prove in Appendix \ref{supp:distilation_guarantees}.

\begin{theorem} Let $\x$ be a $k$-sparse signal under $\D_0$, i.e., there exists $\gama\in \mathbb R^p$ with $\|\gama\|_0\le k$ such that $\x = \D_0\gama$, and let $\hat \D$ be defined as above. Then,
$\argmax_{i} \abs{\x^T \hat \D_i} \in [m]$
as long as \[k \le \frac{1 - \frac{\epsilon}{2} + \sqrt{\epsilon} + \mu(\D_0)}{\mu(\D_0) + \sqrt{\epsilon} + \mu( \D_0,\A)}.\]
\label{thm:guaratee-pruning}
\end{theorem}

Note that, on one hand, if the distance $\epsilon = 0$ and we replace $\mu(\D_0,\A)$ with $\mu(\D_0)$, our condition can be compared to the traditional incoherence condition for OMP that requires $k < \frac{1}{2}(1 + \frac{1}{\mu(\D_0)})$. As shown by the results in \autoref{fig:Exper_3}, we indeed observe that the similarity in the un-related atoms to those in $\D_0$ is quite low, i.e, $\mu(\D_0,\A)$ is very small. Then, in this case (with $\epsilon=0$) our condition is milder than the one for OMP, leading to relaxed and improved guarantees. This is natural, since we must only select atoms in $\hat\D$ that belong to $\hat\D_0$ -- as opposed to demanding the recovery of \emph{the} correct atoms within it. On the other hand, $\A$ itself is allowed to be coherent, even with repeated atoms, as our condition only requires $\mu(\D_0,\A)$ to be small. Lastly, the result above is more general in that we allow for $\epsilon>0$, which better reflects the empirical behavior reflected in \autoref{fig:Exper_3}.

\begin{figure}
    \centering
    \includegraphics[width=\textwidth,trim=60 40 60 40]{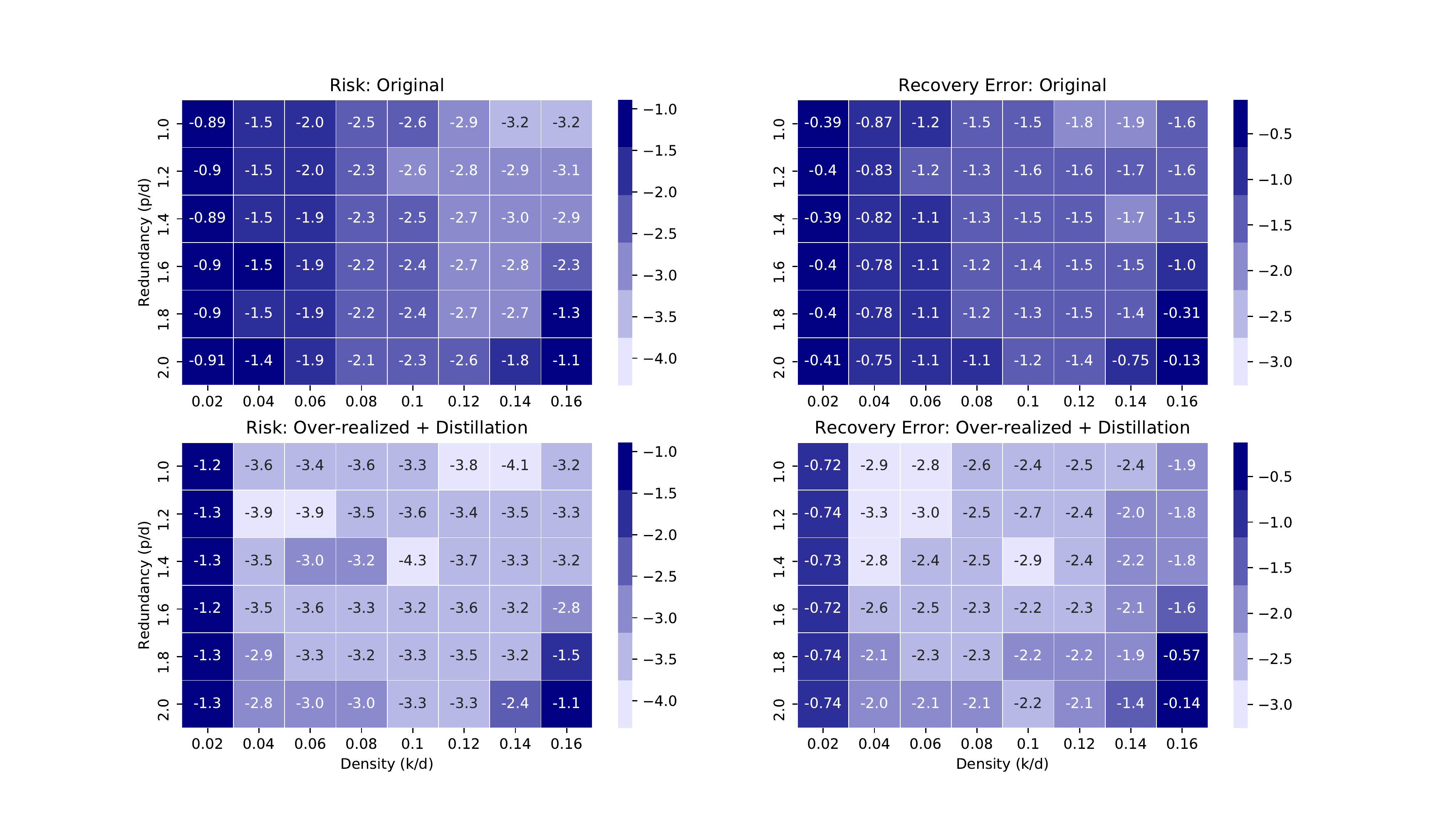}
    \caption{ Risk and Dictionary error ($\log_{10}$ thereof, lower is better) of the estimate provided by \emph{traditional} dictionary learning (i.e. $\hat\D\in\mathcal D_p$) and that resulting from the proposed over-realized approach (i.e. $\hat\D\in\mathcal D_{p'}$) followed by distillation to the original size, over a number of parameters (sparsity, dimension and redundancy).}
    \label{fig:large_scale_OMP} 
\end{figure}

\subsection{Generalization to different model parameters and algorithms}

Thus far we have employed the same experimental setting (dimension, dictionary size and sparsity) for all the above examples for simplicity. However, the reported findings are general and hold for a variety of parameters and algorithms. We now demonstrate this in \autoref{fig:large_scale_OMP} where we report the risk and dictionary error for the estimates produced by learning a dictionary (with ODL+OMP) in the traditional setting (i.e., $\hat\D\in\mathcal D_p$) and that produced by searching over a larger set (i.e., $\hat\D\in\mathcal D_{p'}$, with $p'>p$) followed by our distillation strategy. In this way, all reported measures are computed on estimates of the same size as the original model. Note that an important improvement in risk, but most importantly in dictionary recovery, is observed across a wide range of parameters.
Moreover, the phenomenon is general not just across different model parameters but also to different learning algorithms and regularization functions $g$. In \autoref{supp:numerical_results} we show that similar behaviour (albeit less pronounced) can be obtained by employing: (i) the ODL method from \citep{mairal2010online} with an $\ell_1$ regularizer, i.e. employing Lasso for sparse coding, and (ii) the batch algorithm K-SVD \citep{aharon2006ksvd}.

\begin{figure}[t]
\centering
\subcaptionbox{\label{fig:Noise}}
{\includegraphics[width = .48\textwidth]{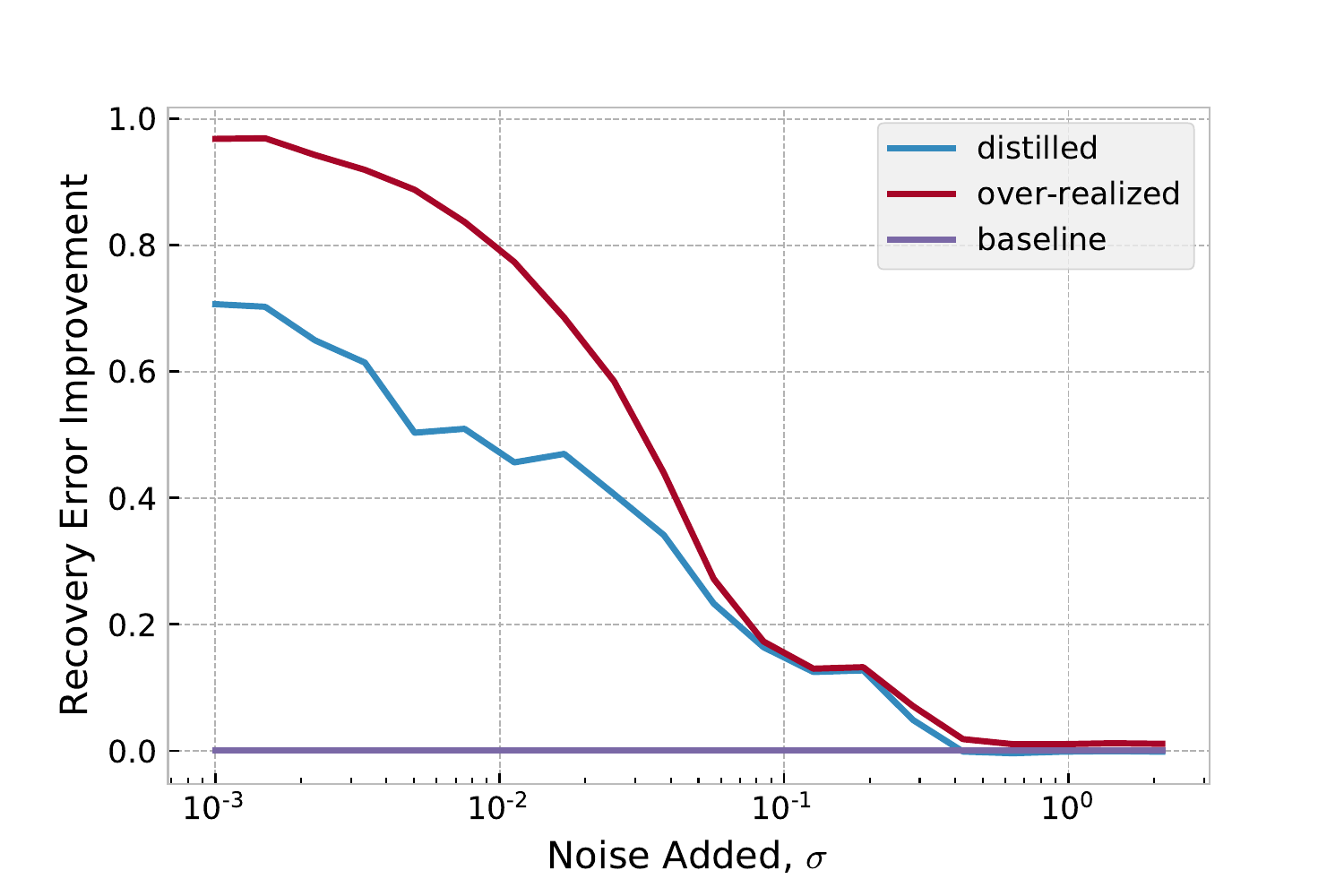}}
\subcaptionbox{\label{fig:Transition}}
{\includegraphics[width = .4\textwidth]{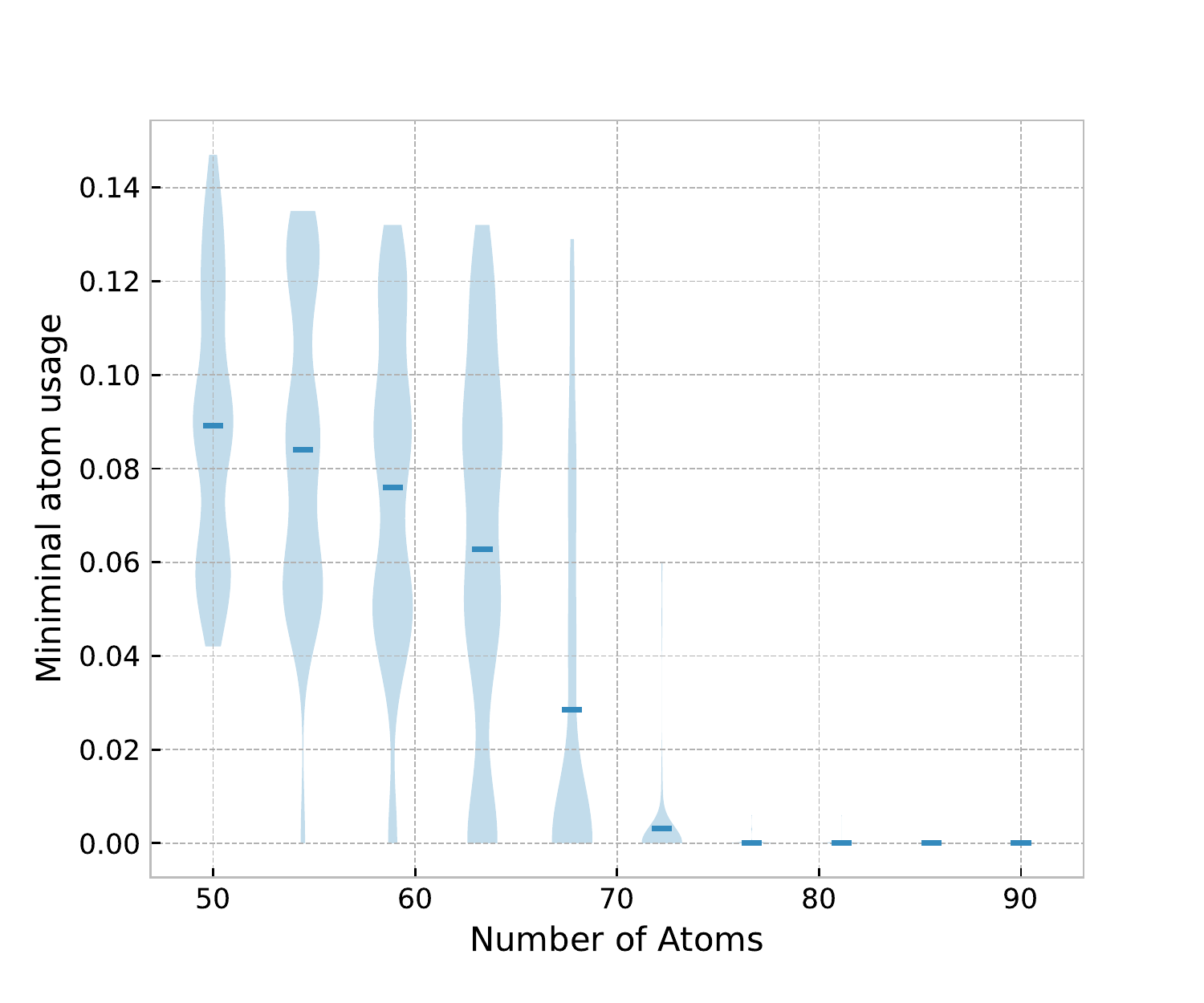}}
\caption{ (a) The improvement in dictionary recovery relative to that of traditional dictionary learning (i.e. searching over dictionaries of the same size) for the over-realized case and our distillation procedure, as a function of added Gaussian noise in the training data. (b) Minimal number of atom usage in the estimated dictionary as a function of the model size, for different dimensions and sparsity levels, exhibiting a phase transition at the size of the ground-truth model (70).  }
\label{fig:Exper_5}
\end{figure}

{On a different note, we have considered the noiseless setting throughout; i.e. each sample $\x$ can be exactly expressed as $\x=\D_0\gama$. In more realistic cases, samples contain measurement noise, or model deviations, which can be modelled by assuming that $\x=\D_0\gama + \v$, where $\v$ is a nuisance vector. While the thorough study of this setting is out of the scope of this work, we will now show empirically that the benefit of over-realization is robust to noise contamination. To this end, and in a similar manner to the above experiments, we contaminate the samples with noise $\v$ sampled from a normal distribution with covariance $\sigma^2\mathbf{I}$. We then measure the risk and recovery error achieved by traditional dictionary learning (i.e., $p'=p$), by the over-realization approach and by our proposed distillation procedure. These quantities are reported in \autoref{fig:Noise} as relative (normalized) improvement over the traditional setting\footnote{More precisely, the quantity measured is $\left(d(\D_0,\hat\D_p)-d(\D_0,\hat\D_{p'})\right)/d(\D_0,\hat\D_p)$ where $\hat\D_p$ is the estimate found by traditional dictionary learning ($p=p'$) and $\hat\D_{p'}$ denotes the estimate found in the over-realized setting, where $p'>p$. The improvement for the distilled version of the estimate is computed analogously.}. As one can see, the benefits of searching over larger model deteriorates smoothly with increasing noise, both for the general over-realized model as well as for our practical distillation approach.} 

\section{Final Remarks } 
\label{sec:conclusion}
In this work we showed that learning over-realized dictionaries can be beneficial not just to provide lower training and population risk, but to also improve the recovery of the underlying model. Our characterization of this phenomenon relies on the connection between the recovery error and the expected risk, enabling thus an upper bound to the former in terms of the empirical risk and a generalization gap. Moreover, we showed that an estimate of the original size can be distilled from the over-realized model, consistently improving the recovered dictionary across different model parameters and algorithms.

{At the same time, several questions remain unanswered. It is still unclear what determines the optimal degree of over-realization. Importantly, a complete understanding of the reasons behind the benefits of over-realization is still missing, and is likely to involve an optimization perspective. A natural hypothesis is that the improvement might be due to having a larger number of ``initial guesses'', since a bigger model will provide a larger covering of the space at initialization. As a result, certain initial atoms will be more likely to fall \emph{close} to some of the ground-truth atoms. This does not seem to be the sole responsible factor, however, as repeating the training process with only those atoms found by distillation (from their initialization) deteriorates performance. }

{
On the other hand, we have noted that the optimization problem presents a phase transition of sorts as the model size grows. We demonstrate this by measuring the minimal number of times any atom in the estimated dictionary is used, for increasing number of atoms, and we compute this statistic for different models dimensions and sparsity levels. One can see in \autoref{fig:Transition} that as soon as the number of atoms exceeds the ground truth size (70 in this case), this statistic drastically drops, reflecting the two-type behaviour illustrated in Section \ref{sec:distill}. On the one hand, this can in fact provide a practical way of determining the (unknown) size of the ground-truth model in practice, which might be worth in its own right. On the other, we believe this might reflects a fundamental change in the optimization landscape. In the $p'>p$ setting, the learning problem might become more amenable to practical optimization algorithms, thus finding a better solution. Further research in this direction will enable to characterize the reported results better, and might extend the application of these ideas to other unsupervised machine learning models.}

\bibliographystyle{plainnat}
\bibliography{references}

\begin{thebibliography}{41}
\providecommand{\natexlab}[1]{#1}
\providecommand{\url}[1]{\texttt{#1}}
\expandafter\ifx\csname urlstyle\endcsname\relax
  \providecommand{\doi}[1]{doi: #1}\else
  \providecommand{\doi}{doi: \begingroup \urlstyle{rm}\Url}\fi

\bibitem[Agarwal et~al.(2014)Agarwal, Anandkumar, Jain, Netrapalli, and
  Tandon]{agarwal2014learning}
Alekh Agarwal, Animashree Anandkumar, Prateek Jain, Praneeth Netrapalli, and
  Rashish Tandon.
\newblock Learning sparsely used overcomplete dictionaries.
\newblock In \emph{Conference on Learning Theory}, pages 123--137, 2014.

\bibitem[Agarwal et~al.(2016)Agarwal, Anandkumar, Jain, and
  Netrapalli]{agarwal2016learning}
Alekh Agarwal, Animashree Anandkumar, Prateek Jain, and Praneeth Netrapalli.
\newblock Learning sparsely used overcomplete dictionaries via alternating
  minimization.
\newblock \emph{SIAM Journal on Optimization}, 26\penalty0 (4):\penalty0
  2775--2799, 2016.

\bibitem[Aharon et~al.(2006{\natexlab{a}})Aharon, Elad, and
  Bruckstein]{aharon2006ksvd}
M.~Aharon, M.~Elad, and A.~Bruckstein.
\newblock {K-SVD}: An algorithm for designing overcomplete dictionaries for
  sparse representation.
\newblock \emph{IEEE Transactions on Signal Processing}, 54\penalty0
  (11):\penalty0 4311--4322, 2006{\natexlab{a}}.

\bibitem[Aharon et~al.(2006{\natexlab{b}})Aharon, Elad, and
  Bruckstein]{aharon2006uniqueness}
Michal Aharon, Michael Elad, and Alfred~M Bruckstein.
\newblock On the uniqueness of overcomplete dictionaries, and a practical way
  to retrieve them.
\newblock \emph{Linear algebra and its applications}, 416\penalty0
  (1):\penalty0 48--67, 2006{\natexlab{b}}.

\bibitem[Arora and Risteski(2017)]{arora2017provable}
Sanjeev Arora and Andrej Risteski.
\newblock Provable benefits of representation learning.
\newblock \emph{arXiv preprint arXiv:1706.04601}, 2017.

\bibitem[Arora et~al.(2014{\natexlab{a}})Arora, Bhaskara, Ge, and
  Ma]{arora2014more}
Sanjeev Arora, Aditya Bhaskara, Rong Ge, and Tengyu Ma.
\newblock More algorithms for provable dictionary learning.
\newblock \emph{arXiv preprint arXiv:1401.0579}, 2014{\natexlab{a}}.

\bibitem[Arora et~al.(2014{\natexlab{b}})Arora, Ge, and Moitra]{arora2014new}
Sanjeev Arora, Rong Ge, and Ankur Moitra.
\newblock New algorithms for learning incoherent and overcomplete dictionaries.
\newblock In \emph{Conference on Learning Theory}, pages 779--806,
  2014{\natexlab{b}}.

\bibitem[Arora et~al.(2015)Arora, Ge, Ma, and Moitra]{arora2015simple}
Sanjeev Arora, Rong Ge, Tengyu Ma, and Ankur Moitra.
\newblock Simple, efficient, and neural algorithms for sparse coding.
\newblock 2015.

\bibitem[Belkin et~al.(2019)Belkin, Hsu, Ma, and Mandal]{belkin2019reconciling}
Mikhail Belkin, Daniel Hsu, Siyuan Ma, and Soumik Mandal.
\newblock Reconciling modern machine-learning practice and the classical
  bias--variance trade-off.
\newblock \emph{Proceedings of the National Academy of Sciences}, 116\penalty0
  (32):\penalty0 15849--15854, 2019.

\bibitem[Bengio et~al.(2013)Bengio, Courville, and
  Vincent]{bengio2013representation}
Yoshua Bengio, Aaron Courville, and Pascal Vincent.
\newblock Representation learning: A review and new perspectives.
\newblock \emph{IEEE transactions on pattern analysis and machine
  intelligence}, 35\penalty0 (8):\penalty0 1798--1828, 2013.

\bibitem[Buhai et~al.(2019)Buhai, Halpern, Kim, Risteski, and
  Sontag]{buhai2019benefits}
Rares-Darius Buhai, Yoni Halpern, Yoon Kim, Andrej Risteski, and David Sontag.
\newblock Empirical study of the benefits of overparameterization in learning
  latent variable models.
\newblock \emph{arXiv preprint arXiv:1907.00030}, 2019.

\bibitem[Candes and Tao(2005)]{candes2005decoding}
Emmanuel~J Candes and Terence Tao.
\newblock Decoding by linear programming.
\newblock \emph{IEEE transactions on information theory}, 51\penalty0
  (12):\penalty0 4203--4215, 2005.

\bibitem[Dasgupta and Schulman(2007)]{dasgupta2007probabilistic}
Sanjoy Dasgupta and Leonard Schulman.
\newblock A probabilistic analysis of em for mixtures of separated, spherical
  gaussians.
\newblock \emph{Journal of Machine Learning Research}, 8\penalty0
  (Feb):\penalty0 203--226, 2007.

\bibitem[Donoho and Elad(2003)]{donoho2003optimally}
David~L Donoho and Michael Elad.
\newblock Optimally sparse representation in general (nonorthogonal)
  dictionaries via $\ell_1$ minimization.
\newblock \emph{Proceedings of the National Academy of Sciences}, 100\penalty0
  (5):\penalty0 2197--2202, 2003.

\bibitem[Elad(2010)]{elad2010sparse}
Michael Elad.
\newblock \emph{Sparse and redundant representations: from theory to
  applications in signal and image processing}.
\newblock Springer Science \& Business Media, 2010.

\bibitem[Engan et~al.(1999)Engan, Aase, and Husoy]{engan1999method}
Kjersti Engan, Sven~Ole Aase, and J~Hakon Husoy.
\newblock Method of optimal directions for frame design.
\newblock In \emph{1999 IEEE International Conference on Acoustics, Speech, and
  Signal Processing. Proceedings. ICASSP99 (Cat. No. 99CH36258)}, volume~5,
  pages 2443--2446. IEEE, 1999.

\bibitem[Geng and Wright(2014)]{geng2014local}
Quan Geng and John Wright.
\newblock On the local correctness of $\ell_1$-minimization for dictionary
  learning.
\newblock In \emph{2014 IEEE International Symposium on Information Theory},
  pages 3180--3184. IEEE, 2014.

\bibitem[Goldt et~al.(2019)Goldt, Advani, Saxe, Krzakala, and
  Zdeborov{\'a}]{goldt2019dynamics}
Sebastian Goldt, Madhu Advani, Andrew~M Saxe, Florent Krzakala, and Lenka
  Zdeborov{\'a}.
\newblock Dynamics of stochastic gradient descent for two-layer neural networks
  in the teacher-student setup.
\newblock In \emph{Advances in Neural Information Processing Systems}, pages
  6979--6989, 2019.

\bibitem[Gribonval et~al.(2015{\natexlab{a}})Gribonval, Jenatton, and
  Bach]{gribonval2015sparse}
R{\'e}mi Gribonval, Rodolphe Jenatton, and Francis Bach.
\newblock Sparse and spurious: dictionary learning with noise and outliers.
\newblock \emph{IEEE Transactions on Information Theory}, 61\penalty0
  (11):\penalty0 6298--6319, 2015{\natexlab{a}}.

\bibitem[Gribonval et~al.(2015{\natexlab{b}})Gribonval, Jenatton, Bach,
  Kleinsteuber, and Seibert]{gribonval2015sample}
R{\'e}mi Gribonval, Rodolphe Jenatton, Francis Bach, Martin Kleinsteuber, and
  Matthias Seibert.
\newblock Sample complexity of dictionary learning and other matrix
  factorizations.
\newblock \emph{IEEE Transactions on Information Theory}, 61\penalty0
  (6):\penalty0 3469--3486, 2015{\natexlab{b}}.

\bibitem[Jung et~al.(2016)Jung, Eldar, and G{\"o}rtz]{jung2016minimax}
Alexander Jung, Yonina~C Eldar, and Norbert G{\"o}rtz.
\newblock On the minimax risk of dictionary learning.
\newblock \emph{IEEE Transactions on Information Theory}, 62\penalty0
  (3):\penalty0 1501--1515, 2016.

\bibitem[Li et~al.(2016)Li, Liang, and Risteski]{li2016recovery}
Yuanzhi Li, Yingyu Liang, and Andrej Risteski.
\newblock Recovery guarantee of non-negative matrix factorization via
  alternating updates.
\newblock In \emph{Advances in neural information processing systems}, pages
  4987--4995, 2016.

\bibitem[Lloyd(1982)]{lloyd1982least}
Stuart Lloyd.
\newblock Least squares quantization in pcm.
\newblock \emph{IEEE transactions on information theory}, 28\penalty0
  (2):\penalty0 129--137, 1982.

\bibitem[Mairal et~al.(2010)Mairal, Bach, Ponce, and Sapiro]{mairal2010online}
Julien Mairal, Francis Bach, Jean Ponce, and Guillermo Sapiro.
\newblock Online learning for matrix factorization and sparse coding.
\newblock \emph{Journal of Machine Learning Research}, 11\penalty0
  (Jan):\penalty0 19--60, 2010.

\bibitem[Maurer and Pontil(2010)]{maurer2010k}
Andreas Maurer and Massimiliano Pontil.
\newblock $ k $-dimensional coding schemes in hilbert spaces.
\newblock \emph{IEEE Transactions on Information Theory}, 56\penalty0
  (11):\penalty0 5839--5846, 2010.

\bibitem[Mei and Montanari(2019)]{mei2019generalization}
Song Mei and Andrea Montanari.
\newblock The generalization error of random features regression: Precise
  asymptotics and double descent curve.
\newblock \emph{arXiv preprint arXiv:1908.05355}, 2019.

\bibitem[Olshausen and Field(1997)]{olshausen1997sparse}
Bruno~A Olshausen and David~J Field.
\newblock Sparse coding with an overcomplete basis set: A strategy employed by
  v1?
\newblock \emph{Vision research}, 37\penalty0 (23):\penalty0 3311--3325, 1997.

\bibitem[Pati et~al.(1993)Pati, Rezaiifar, and
  Krishnaprasad]{pati1993orthogonal}
Yagyensh~Chandra Pati, Ramin Rezaiifar, and Perinkulam~Sambamurthy
  Krishnaprasad.
\newblock Orthogonal matching pursuit: Recursive function approximation with
  applications to wavelet decomposition.
\newblock In \emph{Signals, Systems and Computers, 1993. 1993 Conference Record
  of The Twenty-Seventh Asilomar Conference on}, pages 40--44. IEEE, 1993.

\bibitem[Qu et~al.(2019)Qu, Zhai, Li, Zhang, and Zhu]{qu2019geometric}
Qing Qu, Yuexiang Zhai, Xiao Li, Yuqian Zhang, and Zhihui Zhu.
\newblock Geometric analysis of nonconvex optimization landscapes for
  overcomplete learning.
\newblock In \emph{International Conference on Learning Representations}, 2019.

\bibitem[Schnass(2014)]{schnass2014identifiability}
Karin Schnass.
\newblock On the identifiability of overcomplete dictionaries via the
  minimisation principle underlying k-svd.
\newblock \emph{Applied and Computational Harmonic Analysis}, 37\penalty0
  (3):\penalty0 464--491, 2014.

\bibitem[Seibert(2019)]{seibert2019sample}
Matthias Seibert.
\newblock \emph{Sample Complexity of Representation Learning for Sparse and
  Related Data Models}.
\newblock PhD thesis, Technische Universit{\"a}t M{\"u}nchen, 2019.

\bibitem[Shakeri et~al.(2018)Shakeri, Bajwa, and Sarwate]{shakeri2018minimax}
Zahra Shakeri, Waheed~U Bajwa, and Anand~D Sarwate.
\newblock Minimax lower bounds on dictionary learning for tensor data.
\newblock \emph{IEEE Transactions on Information Theory}, 64\penalty0
  (4):\penalty0 2706--2726, 2018.

\bibitem[Sun et~al.(2015)Sun, Qu, and Wright]{sun2015nonconvex}
Ju~Sun, Qing Qu, and John Wright.
\newblock When are nonconvex problems not scary?
\newblock \emph{arXiv preprint arXiv:1510.06096}, 2015.

\bibitem[Tian(2019)]{tian2019over}
Yuandong Tian.
\newblock Over-parameterization as a catalyst for better generalization of deep
  relu network.
\newblock \emph{arXiv preprint arXiv:1909.13458}, 2019.

\bibitem[Tibshirani(1996)]{tibshirani1996regression}
Robert Tibshirani.
\newblock Regression shrinkage and selection via the lasso.
\newblock \emph{Journal of the Royal Statistical Society: Series B
  (Methodological)}, 58\penalty0 (1):\penalty0 267--288, 1996.

\bibitem[Tillmann(2014)]{tillmann2014computational}
Andreas~M Tillmann.
\newblock On the computational intractability of exact and approximate
  dictionary learning.
\newblock \emph{IEEE Signal Processing Letters}, 22\penalty0 (1):\penalty0
  45--49, 2014.

\bibitem[Tropp(2004)]{tropp2004greed}
Joel~A Tropp.
\newblock Greed is good: Algorithmic results for sparse approximation.
\newblock \emph{IEEE Transactions on Information Theory}, 50\penalty0
  (10):\penalty0 2231--2242, 2004.

\bibitem[Vainsencher et~al.(2011)Vainsencher, Mannor, and
  Bruckstein]{vainsencher2011sample}
Daniel Vainsencher, Shie Mannor, and Alfred~M Bruckstein.
\newblock The sample complexity of dictionary learning.
\newblock \emph{Journal of Machine Learning Research}, 12\penalty0
  (Nov):\penalty0 3259--3281, 2011.

\bibitem[Yang et~al.(2020)Yang, Yu, You, Steinhardt, and
  Ma]{yang2020rethinking}
Zitong Yang, Yaodong Yu, Chong You, Jacob Steinhardt, and Yi~Ma.
\newblock Rethinking bias-variance trade-off for generalization of neural
  networks.
\newblock \emph{arXiv preprint arXiv:2002.11328}, 2020.

\bibitem[Zhai et~al.(2019)Zhai, Yang, Liao, Wright, and Ma]{zhai2019complete}
Yuexiang Zhai, Zitong Yang, Zhenyu Liao, John Wright, and Yi~Ma.
\newblock Complete dictionary learning via $\ell^4$-norm maximization over the
  orthogonal group.
\newblock \emph{arXiv preprint arXiv:1906.02435}, 2019.

\bibitem[Zhang et~al.(2016)Zhang, Bengio, Hardt, Recht, and
  Vinyals]{zhang2016understanding}
Chiyuan Zhang, Samy Bengio, Moritz Hardt, Benjamin Recht, and Oriol Vinyals.
\newblock Understanding deep learning requires rethinking generalization.
\newblock \emph{arXiv preprint arXiv:1611.03530}, 2016.

\end{thebibliography}

\newpage

\begin{center}
    \Large \textbf{Appendix}
\end{center}
\appendix

\section{Recovery Guarantees}
\label{supp:recovery_guarantees}

\setcounter{section}{3}
\renewcommand{\thesection}{\arabic{section}}
\setcounter{theorem}{0}

\begin{lemma} For a ground-truth dictionary $\D_0\in\mathbb{R}^{d \times p}$ generating samples $\x_i=\D_0\gama_i$, where $\gamma_i$ are $k$-sparse with non-zeros sampled iid from a zero mean and unit variance distribution, and for any estimate $\hat\D\in\mathcal D$, with overwhelming probability, we have that
\begin{equation}
   \frac2k~ \underset{\x}{\mathbb{E}}[f^{[k]}_{\x}({\D})] \leq
d(\hat{\D},\D_0) \leq  \frac{4}{k} \underset{\x}{\mathbb{E}}[ f^{[1]}_{\x}({\D})]  - \frac{2}{k} \zeta_k (k-1).
\end{equation}
where $\zeta_k \coloneqq \max\left\{0 ~, ~ 1-(k-2)\mu(\D) - 2\nu(\hat\D, \D_0)^2 \right\}$.
\end{lemma}

\setcounter{section}{1}
\renewcommand{\thesection}{\Alph{section}}

\begin{proof}
Recall that $\x$ is sampled from distribution $\mathbb{P}_k$ by first sampling its support at $\mathcal{S}$ from a uniform distribution of all possible supports with $k$ elements, followed by sampling the non-zeros of its representation given the support, $\gama_\mathcal{S} \sim \mathcal{P}_k$. These non-zero entries are sampled i.i.d. from a distribution with mean zero and variance of 1. The sample is finally constructed as $\x = \D\gama$. 


\paragraph{Upper bound}
Let us first show the upper bound. Let $S = \text{supp}(\gama)$. Then,
\begin{align}
     f^{[1]}_\x(\hat{\D}) =& \inf_{\alfa: \|\alfa\|_0=1} \frac{1}{2}\| \x - \hat{\D}\alfa\|^2_2 \\
     =& \min_j \min_{\alpha_j} \frac{1}{2}\| \D_S\gama_S - \hat\D_j \alpha_j \|^2_2 \\
     =& \frac{1}{2}\| \D_S\gama_S - \hat\D_\sj \left( \hat\D^T_\sj \D_S \gama_S \right)  \|^2_2,
\end{align}
where the last inequality follows by solving for the optimal $\alpha^*_j = \hat\D^T_j\x$, and $\sj$ denotes the optimal choice of the atom index, given by (recall atoms are normalized)
\begin{equation}
    \sj = \arg\min_j \| \x - \hat\D_j \alpha^*_j \|^2_2 = \arg\max_j \big| \langle \D_S\gama_S , \hat{\D}_j \rangle \big|.
\end{equation}
See \citep[Section 3.1]{elad2010sparse} for a more detailed derivation. Let us denote by $\D_i$ the closest atom to $\hat\D_\sj$ in $S$; i.e. $i = \arg\min_{k\in S} \min_{c\in\{+1,-1\}} \| \D_k - c \hat{\D}_\sj \|_2$.
Then, expand the expression above as follows
\begin{align}
    2 f^{[1]}_\x(\hat{\D}) =& \| \left(\D_i\gamma_i + \D_{S\bs i}\gama_{S\bs i}\right) - \hat\D_\sj \hat\D^T_\sj \left(\D_i\gamma_i + \D_{S\bs i}\gama_{S\bs i}\right) \|^2_2 \\
    =& \| (\D_i -\hat\D_\sj \hat{\D}_\sj^T\D_i)\gamma_i + (\mathbf{I} - \hat\D_\sj \hat\D^T_\sj) \D_{S\bs i}\gama_{S\bs i}  \|^2_2  \\
    =& \| (\D_i -\hat\D_\sj \hat{\D}_\sj^T\D_i ) \gamma_i \|_2^2 + \|(\mathbf{I} - \hat\D_\sj\hat\D^T_\sj) \D_{S\bs i}\gama_{S\bs i}  \|^2_2 + \dots \\
    &\dots + 2 \big\langle ~ (\D_i -\hat\D_\sj \hat{\D}_\sj^T\D_i ) \gamma_i ~ ,  ~ (\mathbf{I} - \hat\D_\sj\hat\D_\sj^T) \D_{S\bs i}\gama_{S\bs i} ~ \big\rangle \\
    =&~ A_i + B_i + C_i.
\end{align}
Let us now analyze $\underset{\x\sim\mathbb{P}}{\mathbb{E}} ~[ 2 f^{[1]}_\x(\hat{\D}) ] = \underset{\x\sim\mathbb{P}}{\mathbb{E}} ~[ A_i ] + \underset{\x\sim\mathbb{P}}{\mathbb{E}} ~[ B_i ] + \underset{\x\sim\mathbb{P}}{\mathbb{E}} ~[ C_i ]$.

Consider first
\begin{align}
    \underset{\x\sim\mathbb{P}}{\mathbb{E}} ~[ A_i ] &= \underset{\x\sim\mathbb{P}}{\mathbb{E}} ~[ \| (\D_i -\hat\D_\sj \hat{\D}_\sj^T\D_i ) \gamma_i \|^2_2  ] \\
    &= \underset{S}{\mathbb{E}} \left[ ~~\underset{\gama_S}{\mathbb{E}} ~[  \| \D_i -\hat\D_\sj (\hat{\D}_\sj^T\D_i) \|_2^2 \gamma^2_i  \big| S ] \right] \\
    &= \underset{S}{\mathbb{E}} \left[  \| \D_i -\hat\D_\sj (\hat{\D}_\sj^T\D_i) \|_2^2  \right] \\
    &=  \frac{k}{p} \sum_{i=1}^p \| \D_i - \rho_i \hat\D_\sj \|_2^2  ,
\end{align}
where we used the fact that $\mathbb{E}[\gamma_i^2] = 1$ and we defined $\rho_i \coloneqq \hat{\D}_\sj^T\D_i$.

Looking at the third term,
\begin{align}
    \frac{1}{2}\underset{\x\sim\mathbb{P}}{\mathbb{E}} ~[ C_i ] &=  \underset{\x\sim\mathbb{P}}{\mathbb{E}} ~~  \big\langle ~ (\D_i -\hat\D_\sj \hat{\D}_\sj^T\D_i ) \gamma_i ~ ,  ~ (\mathbf{I} - \hat\D_\sj\hat\D^T_\sj) \D_{S\bs i}\gama_{S\bs i} ~ \big\rangle \\
    &= \underset{S}{\mathbb{E}} \left[ \underset{\gama_S}{\mathbb{E}} ~~ \big[  \big\langle ~ (\D_i - \rho_i \hat\D_\sj ) \gamma_i ~ ,  ~ (\mathbf{I} - \hat\D_\sj \hat\D^T_\sj) \D_{S\bs i}\gama_{S\bs i} ~ \big\rangle \big| S \big] \right] \\
    &= \underset{S}{\mathbb{E}} \left[ ~ \sum_{q\in S\bs i} \underset{\gama_S}{\mathbb{E}} ~~ \big[  \big\langle (\D_i - \rho_i \hat\D_\sj )  \gamma_i , (\mathbf{I} - \hat\D_\sj \hat\D^T_\sj) \D_{q}\gamma_{q} \big\rangle \big| S\big] \right] \\
    &= \underset{S}{\mathbb{E}} \left[ ~ \sum_{q\in S\bs i} \underset{\gama_S}{\mathbb{E}} ~~ [  \gamma_i \gamma_q \langle (\D_i -  \rho_i\hat\D_\sj ) , (\mathbf{I} - \hat\D_\sj\hat\D^T_\sj) \D_{q} \rangle \big| S] \right] \\
    &= 0
    \end{align}
because $\mathbb{E}[ \gamma_i \gamma_q ] = \mathbb{E}[ \gamma_i]\mathbb{E}[\gamma_q ] = 0 $, since the variables are independent and of zero mean. Thus, so far we have that
\begin{align}
     \underset{\x\sim\mathbb{P}}{\mathbb{E}} [ f^{[1]}_\x(\hat{\D}) ] =& \frac{k}{2p} \sum_{i=1}^p \| \D_i - \rho_i \hat\D_\sj \|_2^2 + \frac{1}{2}~\underset{\x\sim\mathbb{P}}{\mathbb{E}} ~[ B_i ]. 
\end{align}
First, note that $\underset{\x\sim\mathbb{P}}{\mathbb{E}} ~[ B_i ]>0$.  Consider a tighter lower bound as follows 
\begin{align}
    \underset{\x\sim\mathbb{P}}{\mathbb{E}} ~[ B_i ] =& \underset{\x\sim\mathbb{P}}{\mathbb{E}} ~~ \|(\mathbf{I} - \hat\D_\sj \hat\D^T_\sj) \D_{S\bs i}\gama_{S\bs i}  \|^2_2 \\
    = &  \underset{\x\sim\mathbb{P}}{\mathbb{E}} ~~ \left[ \| \D_{S\bs i}\gama_{S\bs i}  \|^2_2 + \| \hat\D_\sj \hat\D^T_\sj \D_{S\bs i}\gama_{S\bs i}  \|^2_2  - 2 \langle \D_{S\bs i}\gama_{S\bs i}  , \hat\D_\sj \hat\D^T_\sj \D_{S\bs i}\gama_{S\bs i} \rangle \right] \\
    = &  \underset{\x\sim\mathbb{P}}{\mathbb{E}} ~~ \| \D_{S\bs i}\gama_{S\bs i}  \|^2_2 +\underset{\x\sim\mathbb{P}}{\mathbb{E}} \| \hat\D_\sj \hat\D^T_\sj \D_{S\bs i}\gama_{S\bs i}  \|^2_2  - 2 \underset{\x\sim\mathbb{P}}{\mathbb{E}} (\hat\D_\sj^T\D_{S\bs i}\gama_{S\bs i})^2 \\
    \geq &  \underset{\x\sim\mathbb{P}}{\mathbb{E}} ~~ \| \D_{S\bs i}\gama_{S\bs i}  \|^2_2 - 2 \underset{\x\sim\mathbb{P}}{\mathbb{E}} (\hat\D_\sj^T\D_{S\bs i}\gama_{S\bs i})^2 \\
    = &  \underset{\x\sim\mathbb{P}}{\mathbb{E}} ~~ \| \D_{S\bs i}\gama_{S\bs i}  \|^2_2 - 2 \underset{\x\sim\mathbb{P}}{\mathbb{E}} \left( \sum_{k\in S\bs i} \hat\D_\sj^T\D_k \gama_{k}\right)^2 \\ \label{eq:expectation_trick}
    \geq &  \underset{\x\sim\mathbb{P}}{\mathbb{E}} ~~ \| \D_{S\bs i}\gama_{S\bs i}  \|^2_2 - 2 \max_{k\in S\bs i} \left| \hat\D_\sj^T \D_k \right|^2 \underset{\x\sim\mathbb{P}}{\mathbb{E}} \Big( \sum_{k\in S\bs i} \gama_{k}\Big) ^2 \\
    \geq &  \underset{\x\sim\mathbb{P}}{\mathbb{E}} ~~ \| \D_{S\bs i}\gama_{S\bs i}  \|^2_2 - 2 \max_{k\in [p]\bs i} \left| \hat\D_\sj^T \D_k \right|^2 \underset{\x\sim\mathbb{P}}{\mathbb{E}} \Big( \sum_{k\in S\bs i} \gama_{k}\Big) ^2 \\
    \geq &  \underset{\x\sim\mathbb{P}}{\mathbb{E}} ~~ \| \D_{S\bs i}\gama_{S\bs i}  \|^2_2  - 2 \nu^2 (k-1)
    \end{align}
where we used the fact that $\underset{\x\sim\mathbb{P}}{\mathbb{E}} \Big( \sum_{k\in S\bs i} \gama_{k}\Big) ^2 = k-1$ since the variables are independent and have variance of 1. Additionally, we defined $\nu = \max_j \max_{k\in [p]\bs i^*} \left| \hat\D_{j}^T \D_k \right|$, with $i^*=\arg\max_{k\in[p]} \left| \hat\D_{j}^T \D_k  \right|$. In other words, $i^*$ denotes the nearest neighbor in $\D$ for every $\hat{\D}_j$. 
Continuing from above,
\begin{align}
    \underset{\x\sim\mathbb{P}}{\mathbb{E}} ~[ B_i ] \geq &  \underset{\x\sim\mathbb{P}}{\mathbb{E}} ~~ \| \D_{S\bs i}\gama_{S\bs i}  \|^2_2  - 2 \nu^2 (k-1) \\
    \geq & \underset{\x\sim\mathbb{P}}{\mathbb{E}} ~~ (1-\delta_{k-1})  \| \gama_{S\bs i}  \|^2_2  - 2 \nu^2 (k-1) \\
     \geq & (1-(k-2)\mu(\D)) (k-1) - 2 \nu^2 (k-1) \\ 
     = & \max\{ \big[ 1-(k-2)\mu(\D) - 2\nu^2 \big] (k-1) , 0\}
\end{align}
where $\delta_{k-1}$ is the $(k-1)$-RIP constant of $\D$, and we then used the bound with the mutual coherence $\delta_k\leq(k-1)\mu(\D)$. In the last line, we added the condition that $\underset{\x\sim\mathbb{P}}{\mathbb{E}} ~[ B_i ]\geq0$.

Thus, defining $\zeta_k \coloneqq \max\left\{0,\big[ 1-(k-2)\mu(\D) - 2\nu^2 \big]\right\}$, we can write
\begin{align}
    \underset{\x\sim\mathbb{P}}{\mathbb{E}} [ f^{[1]}_\x(\hat{\D}) ] &\geq \frac{k}{2p} \sum_{i=1}^p \| \D_i - \rho_i \hat\D_\sj \|_2^2 + \frac{1}{2}\big[ 1-(k-2)\mu(\D) - 2\nu^2 \big] (k-1) \\
    &\geq \frac{k}{2p} \sum_{i=1}^p \| \D_i - \rho_i \hat\D_\sj \|_2^2 + \frac{1}{2} \zeta_k(k-1).
\end{align}


Finally, recalling the definition of $\rho_i$ (and that the atoms have unit norm) note that 
\[\| \D_i - (\D_i^T\hat\D_\sj) \hat\D_\sj \|_2^2 \geq \frac{1}{2} \min(\| \D_i - \hat\D_\sj \|_2^2,\| \D_i + \hat\D_\sj \|_2^2) = \frac12 d(\D_i,\hat \D_\sj) \]
Recall that $\D_i$ is the closest atom to $\hat\D_\sj$ out of those in the support $S$, and their distance might be equal or larger to the closest atom in $\hat\D$ to $\D_i$; i.e.
\[ 
{d(\D_i,\hat \D_\sj) \ge \min_j d(\D_i,\hat \D_j)}.
\]
Thus,
\begin{align}
     \frac{1}{p} \sum_{i=1}^p \min_j d(\D_i,\hat \D_j) =  d(\D,\hat\D) \leq \frac{4}{k}\underset{\x\sim\mathbb{P}}{\mathbb{E}} [ f^{[1]}_\x(\hat{\D}) ]  - \frac{2}{k}\zeta_k (k-1).
\end{align}

\paragraph{Lower bound}
Let us know focus on the lower bound for $d(\hat\D,\D)$. For any $S$, let $\hat \D_{\hat S}$ contain the atoms from $\hat \D$ that are closest to the ones in $\D_S$, i.e.,
\[
d(\D_{S(i)},\hat \D_{\hat S(i)}) = d(\D_{S(i)}, \hat \D), \ \forall i \le k.
\]
Then,
\begin{align*}
     f^{[k]}_\x(\hat{\D}) =& \inf_{\alfa: \|\alfa\|_0=k} \frac{1}{2}\| \x - \hat{\D}\alfa\|^2_2 \\
     \le &\min_{\alfa_{\hat S}} \frac{1}{2}\| \D_S\gama_S - \hat\D_{\hat S} \alfa_{\hat S} \|^2_2 \\
     =& \frac{1}{2}\| \D_S\gama_S - \hat\D_{\hat S} (\hat\D_{\hat S}^T \hat\D_{\hat S})^{-1} \hat\D_{\hat S}^T\D_S \gama_S  \|^2_2,
\end{align*}
which implies
\begin{align*}
\mathbb{E}_{\gama_S}[f^{[k]}_\x(\hat{\D})] & = \frac{1}{2} \mathbb{E}_{\gama_S}[\| \D_S\gama_S - \hat\D_{\hat S} (\hat\D_{\hat S}^T \hat\D_{\hat S})^{-1} \hat\D_{\hat S} ^T \D_S \gama_S  \|^2_2] \\
& = \frac{1}{2} \sum_{i=1}^k\| \D_{S(i)} - \hat\D_{\hat S} (\hat\D_{\hat S}^T \hat\D_{\hat S})^{-1} \hat\D_{\hat S}^T \D_{S(i)}  \|^2_2\\
& \leq \frac{1}{2} \sum_{i=1}^k\| \D_{S(i)} - \hat \D_{\hat S(i)} \hat \D_{\hat S(i)}^T \D_{S(i)}  \|^2_2\\
&  \leq \frac{1}{2} \sum_{i=1}^k d(\D_{S(i)},\hat \D_{\hat S(i)}) = \frac{1}{2} \sum_{i=1}^k d(\D_{S(i)},\hat \D),
\end{align*}
where the first line utilizes the fact that each entry of $\gama_S$ is i.i.d. with variance $1$, and the third line follows because $\| \D_{S(i)} - \hat\D_{\hat S} (\hat\D_{\hat S}^T \hat\D_{\hat S})^{-1} \hat\D_{\hat S}^T \D_{S(i)}  \|^2_2$ is the projection residual of $\D_{S(i)}$ onto the subspace spanned by $\hat\D_{\hat S}$, which smaller than the one onto a particular column of $\hat\D_{\hat S}$. The last line follows because 
\[
\|\a - \a \a^T \b\|^2 = \|\a\|^2 - (\a^T\b)^2 \le \min\{\|\a\|^2 - 2(\a^T\b) + \|\b\|^2, \|\a\|^2 + 2(\a^T\b) + \|\b\|^2\} = d(\a,\b)
\] 
for any unit norm vectors $\a,\b \in \mathbb R^d$.
Thus, finally, 
\begin{align*}
\mathbb{E}[f^{[k]}_\x(\hat{\D})] &=  \mathbb{E}_{S}\left[\mathbb{E}_{\gama_S}[f^{[k]}_\x(\hat{\D})]|S\right] \\ 
&\le \frac{1}{2} \mathbb{E}_{S}\left[\sum_{i=1}^k d(\D_{S(i)},\hat \D)\right] \\
&= \frac{1}{2} \frac{\binom{p-1}{k-1}}{\binom{p}{k}} \sum_{i=1}^p d(\D_i,\hat \D) =  \frac{1}{2} \frac{\frac{(p-1)!}{(k-1)!(p-k)!}}{\frac{(p)!}{(k)!(p-k)!}} \sum_{i=1}^p d(\D_i,\hat \D) =\frac{1}{2} \frac{k}{p} \sum_{i=1}^p d(\D_i,\hat \D) \\ &
\le\frac{k}{2} d(\D,\hat \D).
\end{align*}

\end{proof}

\section{Pruning Guarantees}
\label{supp:distilation_guarantees}

Let $\D_0 \in \mathbb R^{d\times p}$ and consider, without loss of generality, that $\hat\D = [\hat\D_0, \A] \in \mathbb R^{d\times p'}$ with $\hat \D_0\in \mathbb R^{d\times m}$, with $m\leq p'$, such that $d(\hat 
\D_i^0,\D_0) \le \epsilon$ for all $i\in \{1,\ldots,m\}$,
and $d(\A_j,\D^0) > \epsilon$ for all $j\in \{1,\ldots, p'-m\}$. In other words, $\hat \D_0$ contains all those $m$ atoms that are $\epsilon$-close to those in $\D_0$, while $\A$ contains those that are further away. Additionally, we require that each atom in $\D_0$ has at least one $\epsilon$-neighbor in $\hat \D_0$; i.e. $d( 
\D^0_i, \hat \D_0) \le \epsilon$ for all $i\in \{1,\ldots,p\}$. We allow $m\ge p$ since the over-realized estimate $\hat \D$ may naturally contain several atoms that close to a real one. Also suppose that both $\D_0$ and $\hat \D$ are column-wise normalized for simplicity.
Let us denote by $\mu(\D_0,\A) = \max_{i,j}\left|\langle  \D_i^0, \A_j \rangle\right|$ the mutual coherence between $\D_0$ and $\A$. With these definitions, we have the following result:

\setcounter{section}{4}
\renewcommand{\thesection}{\arabic{section}}
\setcounter{theorem}{0}

\begin{theorem}{0} 
 Let $\x$ be a $k$-sparse signal under $\D_0$, i.e., there exists $\gama\in \mathbb R^p$ with $\|\gama\|_0\le k$ such that $\x = \D_0\gama$. Then,
$\argmax_{k} \abs{\x^T \hat \d_i} \in [m]$
as long as 
\begin{equation}
k \le \frac{1 - \frac{\epsilon}{2} + \sqrt{\epsilon} + \mu(\D_0)}{\mu(\D_0) + \sqrt{\epsilon} + \mu( \D_0,\A)}.
\label{eq:omp-prun-cond}
\end{equation}
\end{theorem}

\setcounter{section}{2}
\renewcommand{\thesection}{\Alph{section}}

\begin{proof}[Proof of \Cref{thm:guaratee-pruning}] Without of loss generality, we assume that the entries of $\gama$ are placed in the decreasing order of the values $|\gamma_i|$. Recall that we require each atom in $\D_0$ has at least one $\epsilon$-neighbor in $\hat \D_0$; i.e. $d( 
\D^0_i, \hat \D_0) \le \epsilon$ for all $i\in \{1,\ldots,p\}$. For simplicity, we assume $d(\D^0_i,\hat 
\D_i) = \|\D^0_i - \hat 
\D_i\|^2 \le \epsilon$ for all $i\in \{1,\ldots,p\}$, i.e., the $i$-th column of $\hat \D_0$ (or $\hat \D$) is $\epsilon$-close to the $i$-th atom of $\D_0$.

To show the atom that has the largest correlation with $\x$ must be within the first $m$ columns of $\hat\D$, we need to find $i\in[m]$ such tat
\begin{align}
     \left| \x^\top \hat\D_i^0  \right| > \left| \x^\top \A_\ell  \right|, \ \forall \ell.
\label{eq:omp-correct-index-cond}\end{align}

Towards that goal, we choose $i=1$ (as $|\gamma_1|$ is the largest sparse coefficient) to get
\begin{equation}\begin{split}
     \left| \x^\top \hat\D_1^0  \right| = \left| \sum_{i=1}^k \gamma_i (\D_i^0)^\top \hat\D_1^0  \right| &\ge (1-\frac{\epsilon}{2})\left| \gamma_1  \right| - (\mu (\D_0) + \sqrt{\epsilon}) \sum_{i=2}^k \left|  \gamma_i \right| \\
     &\ge \left((1-\frac{\epsilon}{2}) - (k-1)(\mu (\D_0) + \sqrt{\epsilon}) \right)\left| \gamma_1  \right|, 
\end{split}\label{eq:prf-pruning}
\end{equation}
where the first inequality follows because
\[
(\D_1^0)^\top \hat\D_1^0 = 1 - \frac{1}{2}\|\D_1^0 - \hat\D_1^0\|_2^2 \ge 1 - \frac{\epsilon}{2}
\]
and
\[
(\D_i^0)^\top \hat\D_1^0 = (\D_i^0)^\top \D_1^0 + (\D_i^0)^\top(\hat\D_1^0 -\D_1^0) \le \mu(\D_0) + \|\hat\D_1^0 -\D_1^0\|_2 \le \mu(\D_0) + \sqrt{\epsilon}
\]
for all $2\le i\le p$.
On the other hand, we have 
\[
\left| \x^\top \A_\ell  \right| = \left| \sum_{i=1}^k \gamma_i (\D_i^0)^\top \A_\ell  \right| \le \mu(\D_0,\A)  \sum_{i=1}^k \left|  \gamma_i \right| \le k \mu(\D_0,\A) \left| \gamma_1  \right|, \ \forall \ell.
\]
which together with \eqref{eq:prf-pruning} and \eqref{eq:omp-prun-cond} gives \eqref{eq:omp-correct-index-cond}, implying that the chosen element by the first step of OMP must correspond to the one that is close to the correct dictionary, $\D_0$.
\end{proof}

\section{Numerical Results}
\label{supp:numerical_results}

\begin{figure}[h!]
    \centering
    \includegraphics[width=\textwidth,trim=60 20 60 20]{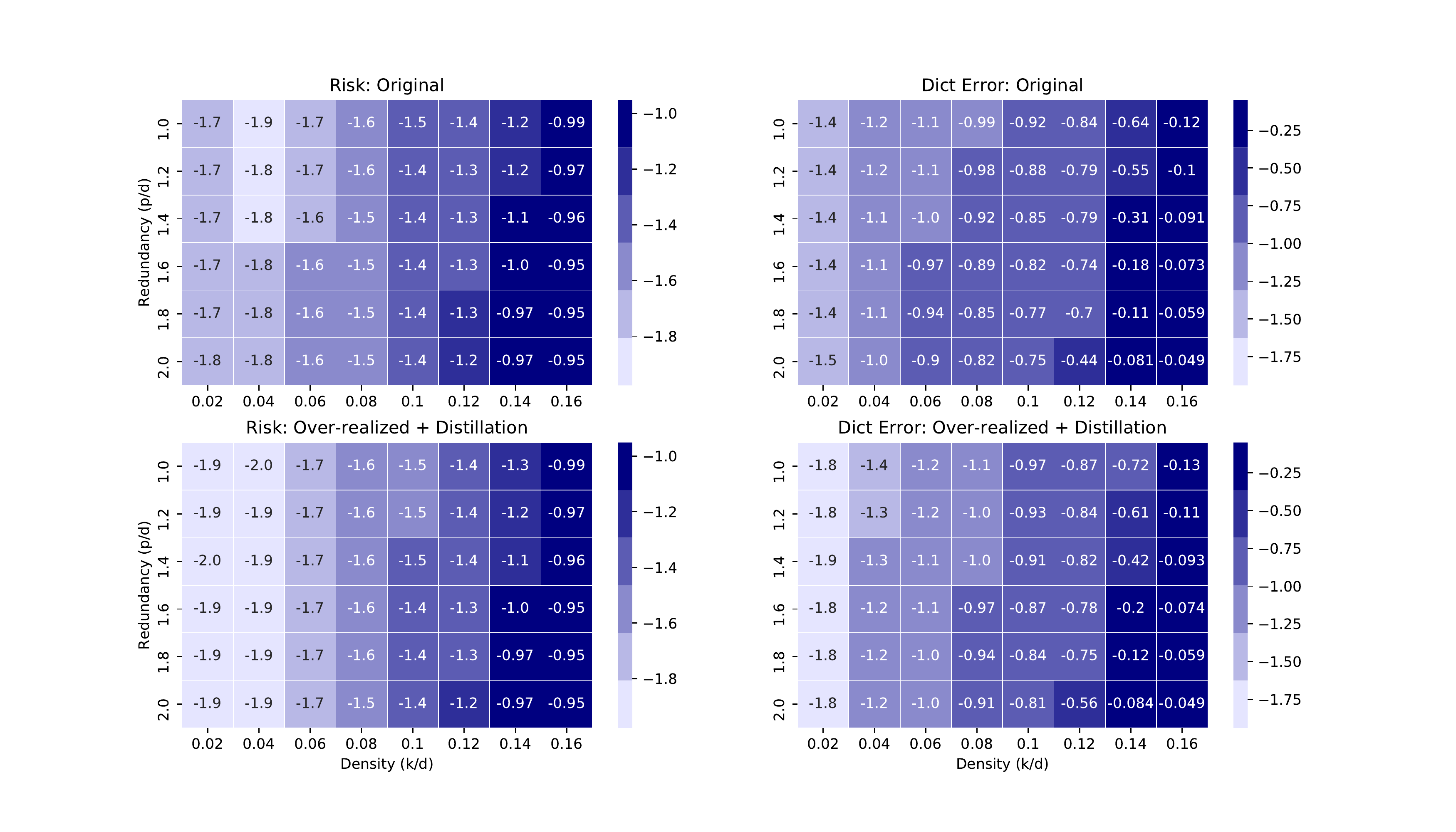}
    \caption{Risk and Dictionary error ($\log_{10}$ thereof, lower is better) of the estimate provided by \emph{traditional} dictionary learning (i.e. $\hat\D\in\mathcal D_p$) and that resulting from the proposed over-realized approach (i.e. $\hat\D\in\mathcal D_{p'}$) followed by distillation to the original size, over a number of parameters (sparsity, dimension and redundancy). Algorithm: ODL+Lasso.}
    \label{fig:large_scale_L1}
\end{figure}

\begin{figure}[h!]
    \centering
    \includegraphics[width=\textwidth,trim=60 20 60 20]{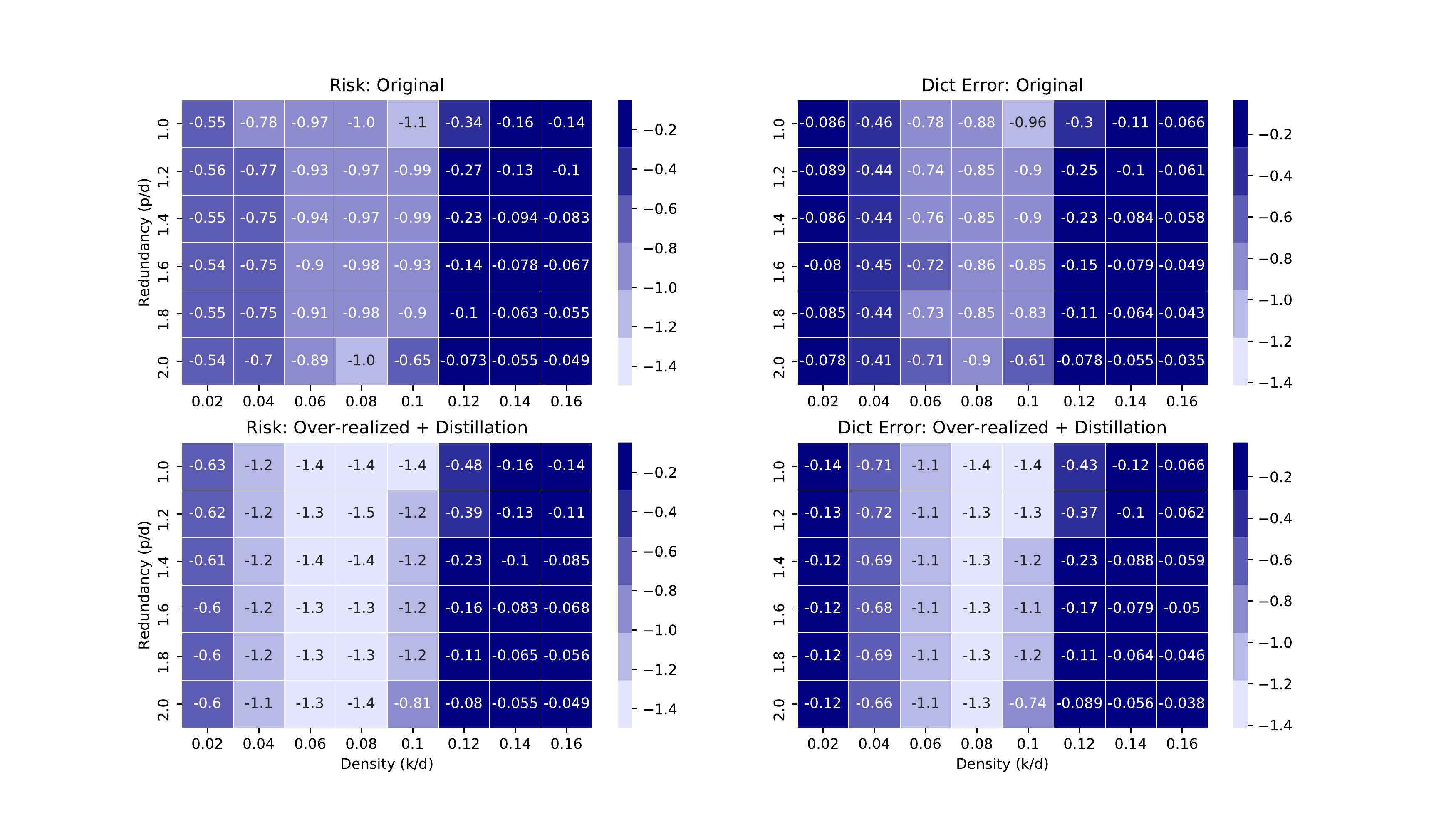}
    \caption{Risk and Dictionary error ($\log_{10}$ thereof, lower is better) of the estimate provided by \emph{traditional} dictionary learning (i.e. $\hat\D\in\mathcal D_p$) and that resulting from the proposed over-realized approach (i.e. $\hat\D\in\mathcal D_{p'}$) followed by distillation to the original size, over a number of parameters (sparsity, dimension and redundancy). Algorithm: K-SVD.}
    \label{fig:large_scale_KSVD}
\end{figure}

\end{document}